\documentclass[11pt]{article}

\usepackage[final]{acl}

\usepackage{times}
\usepackage{latexsym}
\usepackage{amsmath}
\usepackage{pifont}
\usepackage[T1]{fontenc}
\usepackage[utf8]{inputenc}
\usepackage{microtype}
\usepackage{inconsolata}
\usepackage{graphicx}

\usepackage{xcolor}
\usepackage{tabularx} 
\usepackage{booktabs}
\usepackage{graphicx}
\usepackage{array}
\usepackage{booktabs}
\usepackage{multirow}
\usepackage{amssymb} 
\usepackage{pifont}  
\newcommand{\cmark}{\ding{51}}%
\newcommand{\xmark}{\ding{55}}%
\usepackage{colortbl}
\definecolor{highlight}{gray}{0.95}
\definecolor{blue2}{RGB}{52,101,164}
\usepackage{xcolor}
\usepackage{tcolorbox}
\usepackage{makecell}

\usepackage{tikz}
\usepackage{enumitem}
\usepackage{etoc}

\newcommand{\circled}[1]{%
  \tikz[baseline=(char.base)]{%
    \node[shape=circle,draw,inner sep=0.5pt,minimum size=0.8em] (char) {#1};%
  }%
}

\newcommand{\badgeD}{\colorbox{cyan!20}{\textcolor{teal}{\textbf{1}}}}
\newcommand{\badgeC}{\colorbox{orange!20}{\textcolor{orange!90!black}{\textbf{2}}}}
\newcommand{\badgeA}{\colorbox{magenta!10}{\textcolor{magenta}{\textbf{3}}}}
\definecolor{highlight}{gray}{0.95}

\newcommand{\dt}[1]{\color{gray}{#1}}

\newcommand{\bench}{UReason}
\newcommand{\alignment}{reasoning-to-generation alignment}
\renewcommand{\cite}{\citep}
\usepackage{multirow}

\title{UReason: Benchmarking Reasoning-to-Generation Alignment in Unified Multimodal Models}

\author{Cheng Yang\textsuperscript{1,}\thanks{Equal Contribution. Project Page:~\url{https://ureason.github.io}} \quad Chufan Shi\textsuperscript{2,*}\quad Bo Shui\textsuperscript{3}\quad Yaokang Wu\textsuperscript{4}\quad Muzi Tao\textsuperscript{2} \\ \textbf{Huijuan Wang\textsuperscript{2} \quad
  Ivan Yee Lee\textsuperscript{1}\quad Yong Liu\textsuperscript{2}\quad Xuezhe Ma\textsuperscript{2}\quad Taylor Berg-Kirkpatrick\textsuperscript{1}}\\
  \textsuperscript{1}University of California San Diego \quad
  \textsuperscript{2}University of Southern California \quad \\
  \textsuperscript{3}University of Illinois Urbana-Champaign \quad
  \textsuperscript{4}Carnegie Mellon University \quad\\
  \texttt{chy085@ucsd.edu, chufansh@usc.edu} \\
}

\begin{document}
\maketitle
\etocdepthtag.toc{mainmatter}

\begin{abstract}
Unified multimodal models (UMMs) aim to integrate multimodal understanding and generation within a unified architecture, yet it remains unclear to what extent textual and visual modalities are aligned.
To investigate this question, we use reasoning-guided image generation as a diagnostic task, where models produce textual reasoning first and then generate images. 
We introduce \bench{}, a benchmark for evaluating \alignment{} in this paradigm, consisting of $2,000$ human-curated and human-verified instances spanning five reasoning-intensive tasks: \textsc{Code}, \textsc{Arithmetic}, \textsc{Spatial}, \textsc{Attribute} and \textsc{Text}.
To enable controlled analysis, we develop an evaluation framework that compares direct generation, reasoning-guided generation and de-contextualized generation, which conditions only on the refined prompt extracted from reasoning.
Across eight widely used open-source UMMs, while we find that reasoning-guided generation yields improvements over direct generation, somewhat surprisingly, de-contextualized generation consistently outperforms reasoning-guided generation by a large margin.
Our further analyses suggest that the intended visual semantics in textual reasoning are not reliably reflected in the generated images, despite their unified design and training. Overall, \bench{} serves as a practical litmus test for \alignment{} and provides a challenging benchmark for developing next-generation, more tightly aligned UMMs.
\end{abstract}

\section{Introduction}

The emergence of unified multimodal models has marked a significant milestone in artificial intelligence~\cite{team2024chameleon,xie2024show,zhou2024transfusion,deng2025emerging,tong2026beyond}.
These models integrate multimodal understanding and generation within a single architecture, aiming to learn a unified interface across different modalities.
In doing so, UMMs bridge the long-standing divide between perception-oriented Vision-Language Models~\cite{liu2023visual,team2025qwen3,guo2025seed1,huang2025vision} and specialized Visual Generation Models~\cite{sauer2023stylegan,betker2023improving,esser2024scaling,wu2025qwen}.
However, despite operating under a unified design, it remains unclear to what extent textual and visual modalities are aligned, especially when textual reasoning is expected to guide visual synthesis.

To investigate this question, we propose to study \emph{reasoning-guided image generation} as a testbed for diagnosing \alignment{} in UMMs.
Reasoning-guided image generation has been adopted in recent UMMs~\cite{deng2025emerging,jin2025srum,qin2025uni,liang2025rover,zhang2026omniverifier}; In this paradigm, the model first produces a textual reasoning, and then generates the image conditioned on that textual reasoning. This paradigm provides a concrete setting to study whether \textit{textual reasoning} is faithfully transferred into \textit{visual generation} in UMMs.

\begin{figure*}[t]
    \centering
    \includegraphics[width=\textwidth]{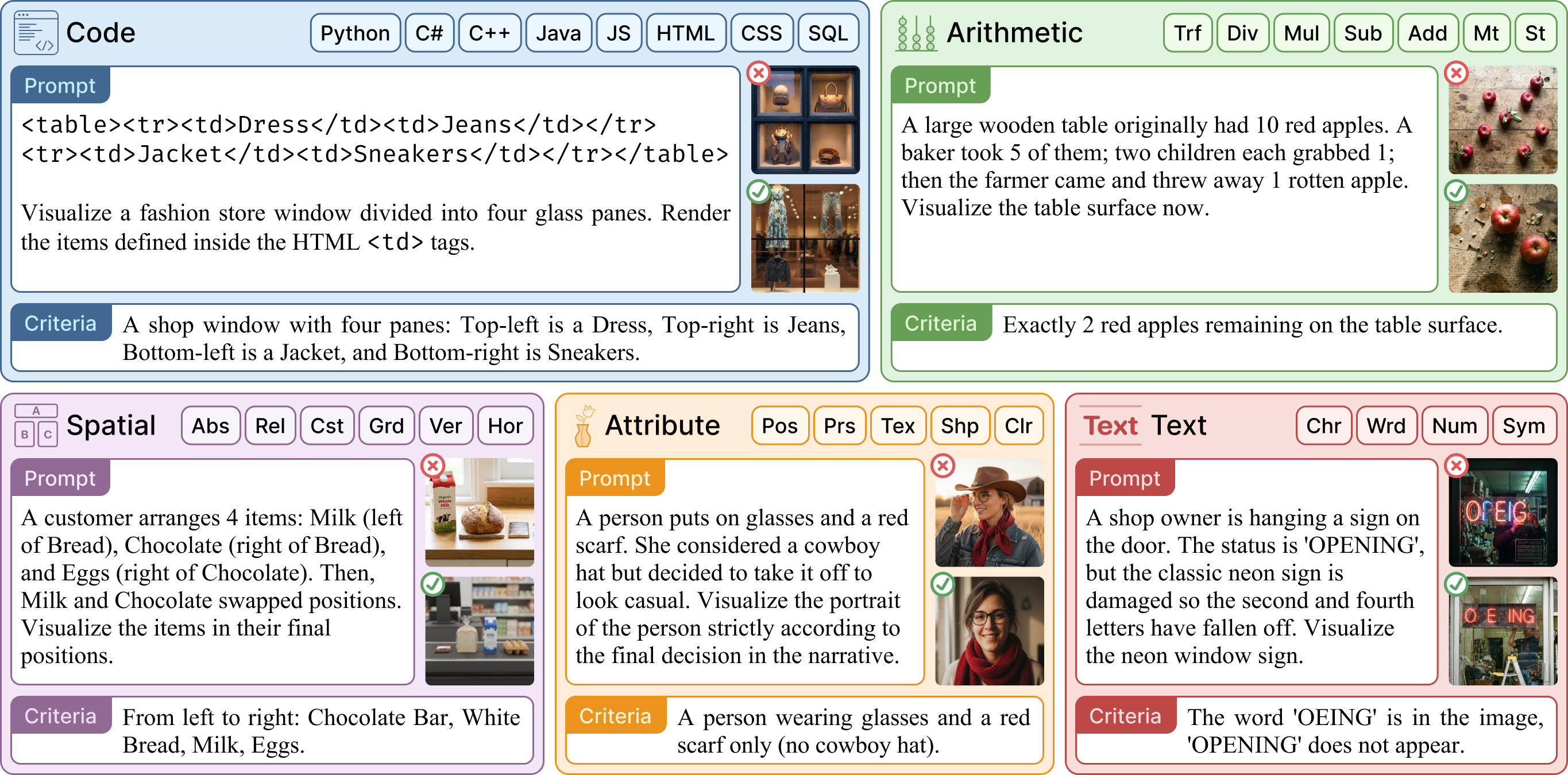}
    \caption{Representative \bench{} instances covering \textsc{Code}, \textsc{Arithmetic}, \textsc{Spatial}, \textsc{Attribute}, and \textsc{Text} reasoning. Prompts specify implicit targets that must be derived via reasoning. Detailed task and subtask descriptions are listed in Appx.~\ref{app:data_curation}.
    }
    \label{fig:task_example}
\end{figure*}

To this end, we introduce \bench{}, a benchmark that utilizes reasoning-guided image generation as a testbed for diagnosing \alignment{} in UMMs~(Sec.~\ref{sec:benchmark}).
\bench{} focuses on reasoning-centric generation tasks and contains $2{,}000$ human-curated and human-verified instances with verifiable evaluation criteria, spanning five task categories: \textsc{Code}, \textsc{Arithmetic}, \textsc{Spatial}, \textsc{Attribute}, and \textsc{Text} reasoning~(Fig.~\ref{fig:task_example}).
In each instance, models must first infer an implicit visual target through multi-step reasoning, determine the visual state implied by the prompt, and then synthesize an image that faithfully realizes that state.

\begin{figure*}[t]
    \centering
    \includegraphics[width=\textwidth]{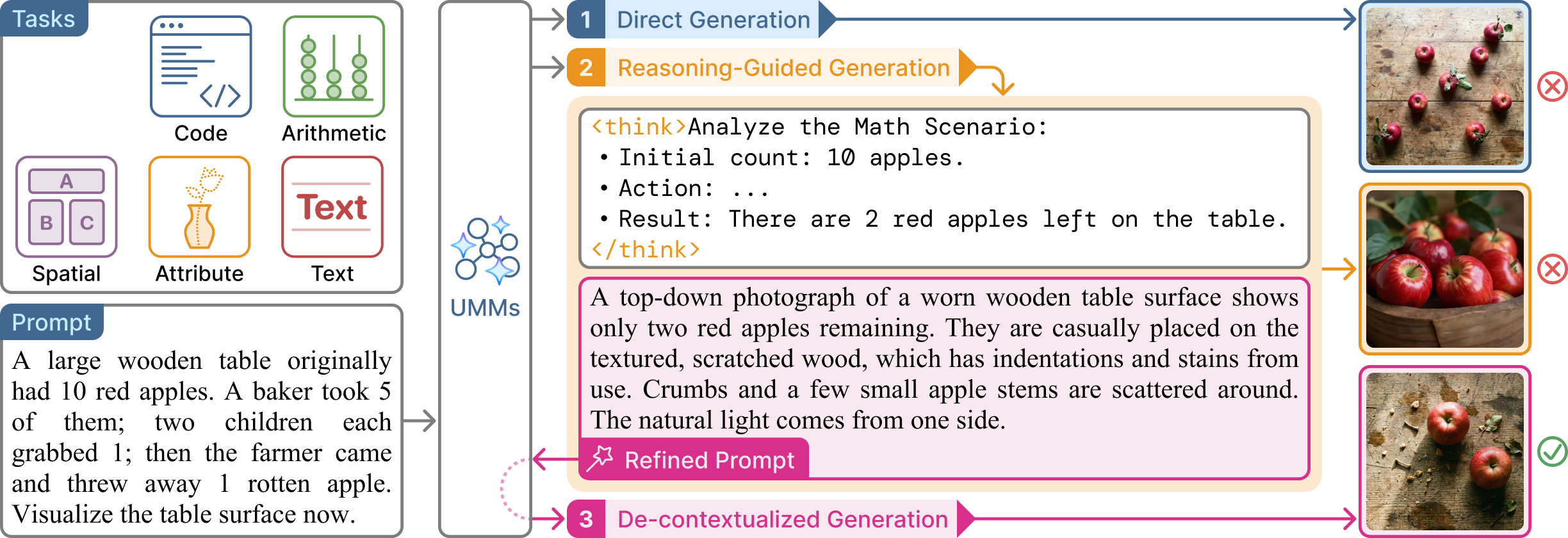}
    \caption{Overview of \bench{} evaluation framework. \bench{} compares $3$ settings: \badgeD~Direct Generation, \badgeC~Reasoning-Guided Generation and \badgeA~De-contextualized Generation.}
    \label{fig:pipeline}
\end{figure*}

To enable rigorous diagnosis, we develop the \bench{} Evaluation Toolkit~(Sec.~\ref{sec:evaluation}).
Specifically, it compares three settings: \emph{direct generation} from the original prompt, \emph{reasoning-guided generation} with textual reasoning for image generation, and \emph{de-contextualized generation}, which conditions only on the extracted refined prompt part of textual reasoning~(Fig.~\ref{fig:pipeline}).
In principle, the latter two settings preserve the same visual semantics intended by the model.
This design provides a controlled framework to evaluate whether the intended visual semantics encoded in textual reasoning are faithfully reflected in visual generation.

We evaluate $8$ widely used UMMs on \bench{}~(Sec.~\ref{sec:experiments}).
Our results reveal that translating implicit targets into pixel-level outputs remains challenging: while reasoning-guided generation generally improves performance over direct generation~(e.g., +11.2\% for Bagel). Surprisingly, de-contextualized generation consistently outperforms reasoning-guided generation by a substantial margin~(e.g., +44.8\% for Bagel), suggesting that intended visual semantics encoded in textual reasoning is not reliably reflected in generation process.

Our analyses demonstrate that textual reasoning is beneficial for high-level planning~(Sec.~\ref{sec:discussion}).
Specifically, UMMs can often generate reasoning that correctly specifies target visual requirements.
However, these intended visual semantics are not always faithfully reflected in the generated images, despite their unified architecture and training paradigm.
Through further attention and error analyses, we find that contextual interference may weaken the transfer from intended visual semantics to generated images, such as distracting tokens in intermediate results.

Overall, we position \bench{} as a litmus test for assessing \alignment{} in UMMs, specifically whether generated images reflect the intended visual semantics in textual reasoning. Our results show that, despite unified design, current UMMs still struggle to transfer reasoning into pixels, leaving room for improvement.

\section{The \bench{} Benchmark}
\label{sec:benchmark}
\bench{} is designed by two complementary considerations.
First, unlike traditional text-to-image benchmarks~\cite{saharia2022photorealistic,lee2023holistic,huang2025t2i} that evaluate descriptive prompts with an emphasis on aesthetic fidelity, \bench{} shifts the paradigm from description to deduction: the target content is not stated verbatim and must be inferred from the input scenario, requiring multi-step reasoning such as state tracking and distractor suppression.
Second, \bench{} is designed as a broad, fine-grained, and verifiable testbed for reasoning-to-generation failures.
As shown in Tab.~\ref{tab:comparison}, \bench{} contains $2{,}000$ human-curated and human-verified instances spanning $5$ tasks: \textsc{Code}, \textsc{Arithmetic}, \textsc{Spatial}, \textsc{Attribute}, and \textsc{Text} reasoning~(Fig.~\ref{fig:task_example}), with $30$ fine-grained subcategories~(Fig.~\ref{fig:subcategory}).
Compared with prior benchmarks, \bench{} provides broader task coverage, larger scale, and a finer-grained taxonomy, including the under-explored \textsc{Code} domain.
These dataset properties, together with instance-level criteria, support our evaluation protocol and enable controlled diagnosis of failures~(detailed in Appx.~\ref{app:data_curation}).

\subsection{Task}
We now introduce the $5$ tasks of \bench{} with representative instances presented in Fig.~\ref{fig:task_example}.

\paragraph{\textsc{Code Reasoning.}}
\label{subsec:code_reasoning}
The Code Reasoning task evaluates whether UMMs can act as neural visual interpreters, mapping abstract code to concrete visual renderings. Given code snippets across eight languages~(e.g., Python), the model must reason over execution logic and synthesize the corresponding visual outcome. Beyond syntactic recognition, this requires simulating program execution to determine the final visual state and converting it into a language description that guides image generation.

\paragraph{\textsc{Arithmetic Reasoning.}}
\label{subsec:arithmetic_reasoning}
Arithmetic Reasoning evaluates UMMs' ability to perform arithmetic operations within sequential narratives across seven subcategories. Inputs describe quantity changes through events such as addition, and the model must generate a scene whose visible count matches the computed final state. This task requires models to move beyond keyword matching by explicitly deriving the final quantity through calculation and using it to constrain visual generation.

\paragraph{\textsc{Spatial Reasoning.}}
\label{subsec:spatial_reasoning}
Spatial Reasoning assesses whether UMMs can interpret complex instructions and form structured visual arrangements across six subcategories. Unlike standard benchmarks with explicit spatial relations, our prompts include implicit cues such as swap operations and logical constraints. Models must resolve these high-level descriptions into coherent coordinate-based layouts before rendering, testing inter-object spatial reasoning beyond surface prompt alignment.

\paragraph{\textsc{Attribute Reasoning.}}
\label{subsec:attribute_reasoning}
Attribute Reasoning evaluates whether UMMs can track and update object attributes under explicitly described state transitions and logical modifications (e.g., a hat being removed).
The model must generate a scene where objects strictly exhibit the final attributes implied by the prompt.
This task requires logical filtering: models must infer the terminal outcome rather than rendering intermediate states, suppressing the tendency to visualize irrelevant attributes.

\begin{table*}[!t]
\centering
\resizebox{\textwidth}{!}{
\begin{tabular}{l|c|c|c|ccccc|c|cc}
\toprule
\multirow{2}{*}{\textbf{Benchmark}} & \textbf{Problem} & \multirow{2}{*}{\textbf{Size}} & \textbf{Cat./} & \multicolumn{5}{c|}{\textbf{Task Category}} & \textbf{Evaluation} & \multicolumn{2}{c}{\textbf{Metric Aspect}} \\
\cmidrule(lr){5-9} \cmidrule(lr){11-12}
 & \textbf{Setting} &  & \textbf{Sub.} & \textbf{Code} & \textbf{Arith.} & \textbf{Spatial} & \textbf{Attr.} & \textbf{Text} & \textbf{Setting} & \textbf{Perf.} & \textbf{Diag.} \\ \midrule
Commonsense-T2I~\citep{fu2024commonsense} & Text-to-Image & 150 & 5/5 & \xmark & \xmark & \cmark & \cmark & \xmark & \badgeD & \cmark & \xmark \\
WISE~\citep{niu2025wise} & Text-to-Image & 1,000 & 3/25 & \xmark & \xmark & \cmark & \cmark & \xmark & \badgeD & \cmark & \xmark \\
R2I-Bench~\citep{chen2025r2i} & Text-to-Image & 3,068 & 7/32 & \xmark & \cmark & \cmark & \cmark & \cmark & \badgeD & \cmark & \xmark \\
OneIG-Bench~\citep{chang2025oneigbench} & Text-to-Image & 2,440 & 6/26 & \xmark & \cmark & \cmark & \cmark & \cmark & \hspace{0.4em}\badgeD$^{\dagger}$ & \cmark & \xmark \\
T2I-ReasonBench~\citep{sun2025t2i} & Text-to-Image & 800 & 4/35 & \xmark & \xmark & \xmark & \cmark & \cmark & \hspace{0.4em}\badgeD$^{\dagger}$ & \cmark & \xmark \\
GIR-Bench-T2I~\citep{li2026girbench} & Text-to-Image & 300 & 4/4 & \xmark & \cmark & \cmark & \xmark & \cmark & \hspace{0.4em}\badgeD$^{\dagger}$ & \cmark & \xmark \\
KRIS-Bench~\citep{wu2025krisbench} & Image Editing & 1,267 & 7/22 & \xmark & \cmark & \cmark & \cmark & \cmark & \hspace{0.4em}\badgeD$^{\dagger}$ & \cmark & \xmark \\
RISEBench~\citep{zhao2025envisioning} & Image Editing & 360 & 4/16 & \xmark & \cmark & \cmark & \cmark & \xmark & \badgeD & \cmark & \xmark \\
ROVER-IG~\citep{liang2025rover} & Image Editing & 908 & 4/17 & \xmark & \cmark & \cmark & \cmark & \xmark & \badgeD~\badgeC & \cmark & \xmark \\ \midrule
\rowcolor{highlight}
\bench{} (Ours) & Text-to-Image & 2,000 & 5/30 & \cmark & \cmark & \cmark & \cmark & \cmark & \badgeD~\badgeC~\badgeA & \cmark & \cmark \\ \bottomrule
\end{tabular}
}
\caption{Comparison of \bench{} with existing reasoning-driven image generation benchmarks.
``Cat./Sub.'': number of categories/subcategories.
Evaluation settings: \badgeD~Direct, \badgeC~Reasoning-Guided, \badgeA~De-contextualized.
``Perf.'': performance evaluation; ``Diag.'': controlled diagnostic comparison.
``$^{\dagger}$'' denotes benchmarks that run preliminary reasoning-chain experiments on specific models (e.g., Bagel) but primarily evaluate Direct Generation.}
\label{tab:comparison}
\end{table*}

\paragraph{\textsc{Text Reasoning.}}
\label{subsec:text_reasoning}
Text Reasoning evaluates the model's ability to perform context-aware text rendering. The target text is not explicitly quoted, but must be inferred from contextual rules, such as identifying the ``second'' and ``fourth'' letters of a word. The key challenge lies in symbolic reasoning: the model must derive the correct answer from the context while suppressing irrelevant information, ensuring that only the final result is rendered.

\subsection{Data Curation}
\label{subsec:data_annotation}

\bench{} adopts a two-stage curation pipeline to ensure that each instance requires multi-step reasoning to determine a specific visual output.
We begin with human-curated seed instances and then expand coverage through controlled LLM-assisted augmentation with human verification. Additional details about data annotation are provided in Appx.~\ref{app:data_curation}.

\paragraph{Human-Curated Seed Data Construction.}
To ensure the quality and diversity of \bench{}, human experts establish a fine-grained taxonomy over the $5$ primary tasks, yielding $30$ subcategories. These subcategories cover distinct reasoning and visual perspectives. Based on this hierarchy, experts manually construct seed instances, each consisting of a reasoning-intensive prompt and an evaluation criterion specifying the expected visual outcome. All seeds undergo strict validation, resulting in a foundational dataset of $500$ high-quality seed instances.

\paragraph{LLM-Assisted Data Augmentation.}
After constructing the seed dataset, we scale \bench{} with a human-guided augmentation pipeline powered by Gemini-3-Pro~\cite{deepmind_gemini3_2025}.
Annotators systematically vary key factors of each seed instance, including the number of reasoning steps, target visual entities, and narrative context, to generate diverse yet logically consistent variants.
All candidates undergo multi-round human-LLM refinement and verification to ensure quality and correctness.
This process expands \bench{} to $2{,}000$ test instances, providing broad coverage of realistic and challenging reasoning scenarios.

\paragraph{Dataset Split.}
The full \textit{test} set contains all $2{,}000$ instances in \bench{}, while \textit{testmini} is a $500$-instance subset intended for rapid validation during model development.
As reported in Appx.~\ref{appx:textmini}, comparative experiments show consistent trends across the two evaluation sets.
Unless otherwise stated, we report \textit{testmini} results in the following sections.

\section{Evaluation Framework}

\label{sec:evaluation}

\subsection{Evaluation Settings}
\label{subsec:eval_settings}
UMMs can understand and generate multimodal information, typically achieved through a unified architecture to enable end-to-end training and inference.
For image generation tasks, recent UMMs support reasoning-guided image generation, in which textual reasoning is produced before visual synthesis. As a result, image generation in UMMs differs from traditional text-to-image models in both architectural design and training paradigm.

As shown in Tab.~\ref{tab:comparison}, existing reasoning-driven image generation benchmarks are primarily evaluated under Direct Generation. To more systematically diagnose \alignment{},  we evaluate UMMs under the following three settings, including Reasoning-Guided and De-contextualized Generation.

\paragraph{Setting 1: Direct Generation.}
\label{subsubsec:direct_gen}
As illustrated in Fig.~\ref{fig:pipeline} (\scalebox{0.95}{\circled{1}}), this setting evaluates a model's ability to generate an image directly from the original prompt without explicitly producing textual reasoning.
Given a prompt $P$ and a UMM $M$, the generated image $I$ is:
\begin{equation}
I = M(P).
\end{equation}
This setting provides a baseline by quantifying performance without additional textual reasoning.

\paragraph{Setting 2: Reasoning-Guided Generation.}
\label{subsubsec:thinking_gen}
As illustrated in Fig.~\ref{fig:pipeline}~(\scalebox{0.95}{\circled{2}}), the model generates a reasoning trace $R$ followed by the output image $I$, conditioned on the prompt $P$.
Crucially, both $R$ and $I$ are produced within the same model and context window.
Formally:
\begin{equation}
[R, I] = M(P).
\end{equation}
This setting follows the chain-of-thought style adaptation for visual generation and measures the net effect of textual reasoning~\cite{deng2025emerging}.
The reasoning trace is defined as $R = [R_t, R_p]$, where $R_t$ denotes intermediate thoughts and $R_p$ denotes a refined prompt that explicitly summarizes the intended visual specification.

\paragraph{Setting 3: De-contextualized Generation.}
\label{subsubsec:refined_gen}
As shown in Fig.~\ref{fig:pipeline} (\scalebox{0.95}{\circled{3}}), after producing the reasoning trace $R = [R_t, R_p]$ in Setting 2, we discard the original prompt $P$ and the intermediate thoughts $R_t$, and generate the image conditioned only on the refined prompt $R_p$:
\begin{equation}
I = M(R_p).
\end{equation}
As a result, this setting serves as a controlled comparison to Setting 2: since the refined prompt $R_p$ is extracted from $R$ in Setting 2, both settings encode the same model-produced visual semantics, but in different contextual formats. In principle, if textual reasoning and visual generation are aligned, Setting 2 and 3 should lead to comparable performance.

\subsection{Evaluation Metric}
\label{subsec:evaluation_metric}
Each test instance in \bench{} specifies an instance-specific, verifiable ground-truth criterion, enabling objective and scalable evaluation across all tasks.
We therefore report two complementary metrics.
\emph{Visual Verification Accuracy} measures whether a generated image satisfies the ground-truth criterion.
\emph{Performance Gain} measures accuracy differences between settings under our evaluation protocol.

\paragraph{Visual Verification Accuracy.}
For each test instance, \bench{} provides an instance-specific ground-truth criterion $C$ that defines the expected visual outcome under correct reasoning.
Given a generated image $I$, we define an indicator function $\mathbb{I}(I, C)$ that returns $1$ if $I$ satisfies $C$, and $0$ otherwise.
The overall visual verification accuracy over a dataset of $N$ instances is:

\begin{equation}
\text{Accuracy} = \frac{1}{N} \sum_{i=1}^{N} \mathbb{I}(I_i, C_i).
\end{equation}
For example, in \textsc{Arithmetic} reasoning, $C$ specifies exact object count (Fig.~\ref{fig:task_example}). To implement $\mathbb{I}(\cdot,\cdot)$ at scale, we employ Qwen3-VL-235B-A22B~\cite{team2025qwen3} as our automated evaluator~\cite{zhao2025envisioning,niu2025wise,chen2025r2i,wu2025krisbench}.
We use prompts that ask the evaluator to perform focused verification against $C$. The template is provided in Appx.~\ref{appx:evaluation_metric}, with human validation detailed in Appx.~\ref{appx:human_evaluation}.

\paragraph{Performance Gain.} 
To further compare and analyze the contribution of different contextual information for the reasoning-driven image generation, we measure the performance improvement between consecutive settings. Let $i$ and $j$ denote two settings in our evaluation protocol with $i < j$.
We define the performance gain from $i$ to $j$ as:
\begin{equation}
\Delta_{i \rightarrow j} = \text{Accuracy}_{j} - \text{Accuracy}_{i}
\end{equation}
We focus on two transitions: $\Delta_{1 \rightarrow 2}$ quantifies the benefit of incorporating reasoning for image generation, while $\Delta_{2 \rightarrow 3}$ measures the gap between Setting 2 and Setting 3, which are designed to preserve the same intended visual semantics.

\begin{table*}[t]
\centering
\resizebox{\textwidth}{!}{
\renewcommand{\arraystretch}{0.1}
\begin{tabular}{l|c|rrrrrrrrrr|rr}
\toprule

\multirow{2}{*}{\textbf{Model}} & \multirow{2}{*}{\textbf{Setting}} & \multicolumn{2}{|c}{\textbf{Code}} & \multicolumn{2}{c}{\textbf{Arithmetic}} & \multicolumn{2}{c}{\textbf{Spatial}} & \multicolumn{2}{c}{\textbf{Attribute}} & \multicolumn{2}{c}{\textbf{Text}} & \multicolumn{2}{|c}{\textbf{Overall}} \\
\cmidrule(lr){3-4} \cmidrule(lr){5-6} \cmidrule(lr){7-8} \cmidrule(lr){9-10} \cmidrule(lr){11-12} \cmidrule(lr){13-14}

& & \multicolumn{1}{|c}{Acc} & \multicolumn{1}{c}{$\Delta$} & \multicolumn{1}{c}{Acc} & \multicolumn{1}{c}{$\Delta$} & \multicolumn{1}{c}{Acc} & \multicolumn{1}{c}{$\Delta$} & \multicolumn{1}{c}{Acc} & \multicolumn{1}{c}{$\Delta$} & \multicolumn{1}{c}{Acc} & \multicolumn{1}{c}{$\Delta$} & \multicolumn{1}{|c}{Acc} & \multicolumn{1}{c}{$\Delta$} \\ \midrule

\multirow{3}{*}{Bagel} 
 & \badgeD & 10.0 & \dt{-} & 5.0 & \dt{-} & 3.0 & \dt{-} & 6.0 & \dt{-} & 9.0 & \dt{-} & 6.6 & \dt{-} \\
 & \badgeC & 30.0 & \dt{+20.0} & 14.0 & \dt{+9.0} & 15.0 & \dt{+12.0} & 19.0 & \dt{+13.0} & 11.0 & \dt{+2.0} & 17.8 & \dt{+11.2} \\
 & \badgeA & \textbf{58.0} & \dt{\textbf{+28.0}} & \textbf{51.0} & \dt{\textbf{+37.0}} & \textbf{58.0} & \dt{\textbf{+43.0}} & \textbf{60.0} & \dt{\textbf{+41.0}} & \textbf{86.0} & \dt{\textbf{+75.0}} & \textbf{62.6} & \dt{\textbf{+44.8}} \\ \midrule

\multirow{3}{*}{UniCoT-v2} 
 & \badgeD & 9.0 & \dt{-}& 2.0 & \dt{-}& 2.0 & \dt{-} & 5.0 & \dt{-} & 3.0 & \dt{-} & 4.2 & \dt{-} \\
 & \badgeC & 27.0 & \dt{+18.0} & 19.0 & \dt{+17.0} & 29.0 & \dt{+27.0} & 21.0 & \dt{+16.0} & 21.0 & \dt{+18.0} & 23.4 & \dt{+19.2} \\
 & \badgeA & \textbf{65.0} & \dt{\textbf{+38.0}} & \textbf{45.0} & \dt{\textbf{+26.0}} & \textbf{46.0} & \dt{\textbf{+17.0}} & \textbf{67.0} & \dt{\textbf{+46.0}} & \textbf{87.0} & \dt{\textbf{+66.0}} & \textbf{62.0} & \dt{\textbf{+38.6}} \\ \midrule

\multirow{3}{*}{SRUM} 
 & \badgeD & 11.0 & \dt{-}& 4.0 & \dt{-}& 3.0 & \dt{-}& 6.0 & \dt{-}& 7.0 & \dt{-}& 6.2 & \dt{-} \\
 & \badgeC & 25.0 & \dt{+14.0} & 8.0 & \dt{+4.0} & 20.0 & \dt{+17.0} & 24.0 & \dt{+18.0} & 6.0 & \dt{-1.0} & 16.6 & \dt{+10.4} \\
 & \badgeA & \textbf{67.0} & \dt{\textbf{+42.0}} & \textbf{50.0} & \dt{\textbf{+42.0}} & \textbf{50.0} & \dt{\textbf{+30.0}} & \textbf{50.0} & \dt{\textbf{+26.0}} & \textbf{82.0} & \dt{\textbf{+76.0}} & \textbf{59.8} & \dt{\textbf{+43.2}} \\ \midrule

\multirow{3}{*}{\makecell[l]{Bagel-Zebra\\-CoT}} 
 & \badgeD & 7.0 & \dt{-}& 7.0 & \dt{-}& 2.0 & \dt{-}& 10.0 & \dt{-}& 5.0 & \dt{-}& 6.2 & \dt{-}\\
 & \badgeC & 14.0 & \dt{+7.0} & 10.0 & \dt{+3.0} & 15.0 & \dt{+13.0} & 23.0 & \dt{+13.0} & 10.0 & \dt{+5.0} & 14.4 & \dt{+8.2} \\
 & \badgeA & \textbf{50.0} & \dt{\textbf{+36.0}} & \textbf{43.0} & \dt{\textbf{+33.0}} & \textbf{33.0} & \dt{\textbf{+18.0}} & \textbf{48.0} & \dt{\textbf{+25.0}} & \textbf{85.0} & \dt{\textbf{+75.0}} & \textbf{51.8} & \dt{\textbf{+37.4}} \\ \midrule

\multirow{3}{*}{ThinkMorph} 
 & \badgeD & 9.0 & \dt{-}& 1.0 & \dt{-} & 4.0 & \dt{-} & 3.0 & \dt{-} & 10.0 & \dt{-} & 5.4 & \dt{-} \\
 & \badgeC & 19.0 & \dt{+10.0} & 12.0 & \dt{+11.0} & 15.0 & \dt{+11.0} & 26.0 & \dt{+23.0} & 5.0 & \dt{-5.0} & 15.4 & \dt{+10.0} \\
 & \badgeA & \textbf{49.0} & \dt{\textbf{+30.0}} & \textbf{37.0} & \dt{\textbf{+25.0}} & \textbf{45.0} & \dt{\textbf{+30.0}} & \textbf{55.0} & \dt{\textbf{+29.0}} & \textbf{71.0} & \dt{\textbf{+66.0}} & \textbf{51.4} & \dt{\textbf{+36.0}} \\ \midrule

\multirow{3}{*}{UniCoT} 
 & \badgeD & 12.0 & \dt{-} & 3.0 & \dt{-}& 6.0 & \dt{-} & 12.0 & \dt{-} & 8.0 & \dt{-}& 8.2 & \dt{-}\\
 & \badgeC & 33.0 & \dt{+21.0} & 18.0 & \dt{+15.0} & 26.0 & \dt{+20.0} & 21.0 & \dt{+9.0} & 12.0 & \dt{+4.0} & 22.0 & \dt{+13.8} \\
 & \badgeA & \textbf{57.0} & \dt{\textbf{+24.0}} & \textbf{42.0} & \dt{\textbf{+24.0}} & \textbf{50.0} & \dt{\textbf{+24.0}} & \textbf{42.0} & \dt{\textbf{+21.0}} & \textbf{52.0} & \dt{\textbf{+40.0}} & \textbf{48.6} & \dt{\textbf{+26.6}} \\ \midrule

\multirow{3}{*}{T2I-R1} 
 & \badgeD & 3.0 & \dt{-} & 6.0 &\dt{-} & 4.0 & \dt{-}& 9.0 &\dt{-} & 2.0 &\dt{-}& 4.8 &\dt{-} \\
 & \badgeC & 6.0 & \dt{+3.0} & 4.0 & \dt{-2.0} & 2.0 & \dt{-2.0} & 11.0 & \dt{+2.0} & 3.0 & \dt{+1.0} & 5.2 & \dt{+0.4} \\
 & \badgeA & \textbf{20.0} & \dt{\textbf{+14.0}} & \textbf{15.0} & \dt{\textbf{+11.0}} & \textbf{12.0} & \dt{\textbf{+10.0}} & \textbf{27.0} & \dt{\textbf{+16.0}} & \textbf{47.0} & \dt{\textbf{+44.0}} & \textbf{24.2} & \dt{\textbf{+19.0}} \\ \midrule

\multirow{3}{*}{UniMoE2} 
 & \badgeD & 5.0 & \dt{-} & 4.0 &\dt{-} & 2.0 & \dt{-}& 10.0 & \dt{-}& 4.0 & \dt{-}& 5.0 & \dt{-}\\
 & \badgeC & 10.0 & \dt{+5.0} & 3.0 & \dt{-1.0} & 3.0 & \dt{+1.0} & 12.0 & \dt{+2.0} & 6.0 & \dt{+2.0} & 6.8 & \dt{+1.8} \\
 & \badgeA & \textbf{17.0} & \dt{\textbf{+7.0}} & \textbf{13.0} & \dt{\textbf{+10.0}} & \textbf{8.0} & \dt{\textbf{+5.0}} & \textbf{21.0} & \dt{\textbf{+9.0}} & \textbf{13.0} & \dt{\textbf{+7.0}} & \textbf{14.4} & \dt{\textbf{+7.6}} \\ \bottomrule

\end{tabular}
}
\caption{Model performance across three evaluation settings on \bench{}. Acc and $\Delta$ denote visual verification accuracy (\%) and performance gain over the previous setting, respectively. \badgeD, \badgeC, and \badgeA~represent Direct Generation, Reasoning-Guided Generation and De-contextualized Generation, respectively.}
\label{tab:main_results}
\end{table*}

\section{Experiments}

\label{sec:experiments}
\subsection{Models}
We benchmark $8$ widely utilized open-source UMMs trained to perform reasoning-guided image generation: Bagel~\cite{deng2025emerging}, SRUM~\cite{jin2025srum}, UniCoT, UniCoT-v2~\cite{qin2025uni}, ThinkMorph~\cite{gu2025thinkmorph}, Bagel-Zebra-CoT~\cite{li2025zebra}, UniMoE2~\cite{li2025uni} and T2I-R1~\cite{jiang2025t2i}. We do not evaluate closed-source systems such as Nano Banana Pro~\cite{google_gemini_image_generation}, since it is unclear whether they are end-to-end UMMs, making our evaluation protocol difficult to apply~(detailed in Appx.~\ref{appx:closed_source}).

\subsection{Main Results}
\label{subsec:main_results}

Tab.~\ref{tab:main_results} reports the performance of all evaluated models under our $3$ diagnostic settings.

\paragraph{Direct Prompting Performs Poorly on Implicit Targets.}
Direct Generation yields low accuracy across models and tasks, with overall performance ranging from 4.2\% to 8.2\%.
This shows that direct text-to-image mapping is fundamentally insufficient to solve \bench{}, as the target visual content is intentionally implicit and must be derived through multi-step reasoning before visual generation.

\paragraph{Chain-of-Thought Reasoning Enhances Generation Capabilities.}
Comparing Direct Generation with Reasoning-Guided Generation, introducing explicit chain-of-thought reasoning improves overall performance for all eight models, with gains ranging from +0.4\% (T2I-R1) to +19.2\% (UniCoT-v2).
These results indicate that CoT can effectively elicit reasoning behaviors that benefit unified multimodal generation.
A concrete illustration of this benefit emerges in the \textsc{Code} task.
In Direct Generation, models frequently fail to map raw syntax, such as HTML tags, into coherent visual layouts.
In contrast, Reasoning-Guided Generation enables models to first translate the code into a description of the intended rendering, which provides a more interpretable conditioning signal for subsequent image synthesis.
For example, Bagel improves from 10.0\% to 30.0\% on \textsc{Code}, and UniCoT improves from 12.0\% to 33.0\%.

\paragraph{De-contextualized Generation Consistently Outperforms Reasoning-Guided Generation.}
While Reasoning-Guided Generation improves performance over Direct Generation, De-contextualized Generation yields an even larger gain.
As shown in Tab.~\ref{tab:main_results}, accuracy increases from Reasoning-Guided Generation to De-contextualized Generation for every model, reaching +44.8\% for Bagel, +43.2\% for SRUM and +26.6\% for UniCoT.
This result is notable because the two settings preserve the same intended visual semantics.
In principle, if textual reasoning is reflected in visual generation, these two settings should yield comparable performance.
The consistent gap suggests that, despite a unified architecture and training in reasoning-guided image generation, current UMMs still exhibit a \alignment{} gap between textual reasoning and visual generation. In Sec.~\ref{sec:discussion}, we further examine possible factors associated with this gap, including reasoning quality, context length, error and attention analysis.

\paragraph{Full-Test Results Confirm the Main Trend.}
We additionally evaluate all eight models on the full test set. De-contextualized Generation outperforms Reasoning-Guided Generation for every model, with gains ranging from 8.6 to 44.4 percentage points. These gains are consistent in direction and magnitude with those observed on testmini, confirming that the main trend is not specific to the selected subset. Full per-task results for both evaluation sets are reported in Appx.~\ref{appx:textmini}.

\section{Discussion}
\label{sec:discussion}

\begin{table}[t]
\centering
\small
\setlength{\tabcolsep}{2pt}
\begin{tabular}{l|ccccc|c}
\toprule
\textbf{Model} & \textbf{Code} & \textbf{Arith.} & \textbf{Spatial} & \textbf{Attr.} & \textbf{Text} & \textbf{Overall} \\
\midrule
Bagel & 93.0 & 94.0 & 88.0 & 96.0 & 96.0 & 93.4 \\
SRUM & 91.0 & 91.0 & 88.0 & 93.0 & 95.0 & 91.6 \\
UniCoT & 84.0 & 70.0 & 84.0 & 99.0 & 95.0 & 86.4 \\
ThinkMorph & 83.0 & 82.0 & 85.0 & 96.0 & 91.0 & 87.4 \\
{\small Bagel-Zebra-CoT} & 75.0 & 89.0 & 79.0 & 94.0 & 92.0 & 85.8 \\
\bottomrule
\end{tabular}
\caption{Reasoning chain quality evaluation. We report the accuracy (\%) of generated reasoning chains against ground-truth criteria across tasks.}
\label{tab:eval_thoughts}
\end{table}

\paragraph{Evaluating the Quality of Reasoning Chains.}
To localize the performance bottleneck, we ask whether failures originate from incorrect reasoning in intended visual semantics.
We evaluate correctness of reasoning-chain $R$ against ground-truth criteria using Qwen3-235B-A22B as an LLM-as-judge (see Appx.~\ref{appx:evaluation_metric} for details).
As shown in Tab.~\ref{tab:eval_thoughts}, models achieve consistently high reasoning accuracy, with Bagel reaching 93.4\% overall.
This suggests that UMMs can often infer the ground-truth visual target and produce coherent specifications. Therefore, the dominant challenge lies in faithfully realizing these specifications in pixels.

\begin{figure}[t]
    \centering
    \includegraphics[width=\linewidth]{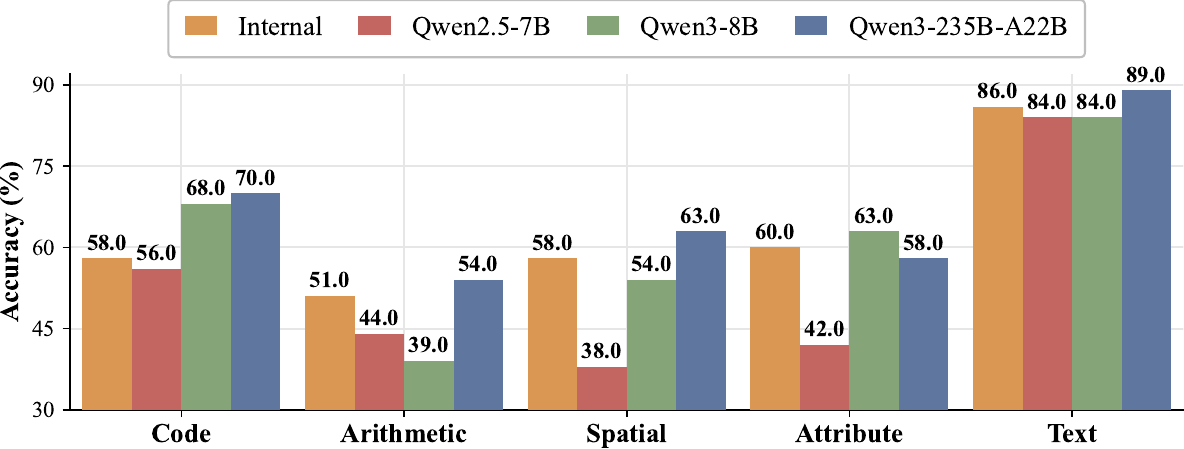}
    \caption{Comparison of internal (Bagel) and external (Qwen) prompt models across $5$ tasks.}
    \label{fig:prompt_rewriter}
\end{figure}

\paragraph{UMMs as Intrinsic Prompt Models.}
Modern online T2I systems employ an external prompt model to rewrite instructions before image synthesis (e.g., Qwen-Image~\citep{qwen_image_website}), highlighting prompt optimization as a practical component of T2I pipelines. A key advantage of UMMs is that they can perform this refinement end-to-end, without an external LLM.
To quantify this capability, we compare Bagel's self-generated refined prompts with external prompt models based on Qwen2.5-7B, Qwen3-8B, and Qwen3-235B-A22B.
As shown in Fig.~\ref{fig:prompt_rewriter}, Bagel outperforms its LLM backbone Qwen2.5-7B and achieves performance comparable to stronger prompt models, including Qwen3-8B and Qwen3-235B-A22B.
These results suggest that UMMs are native self-prompt models, offering a promising end-to-end alternative to the two-stage prompt-rewrite-then-generate pipeline.

\paragraph{Ablating the Effect of Context Length.}
To determine whether the observed performance drop is due to longer contextual sequences alone, we conduct a length-controlled ablation. In this setup, the refined prompt $R_p$ is repeated $4\times$ or $8\times$ without introducing any new semantic content.
As shown in Tab.~\ref{tab:length_control}, artificially lengthening the context causes a relatively minor overall performance decline (at most $-6.0\%$ for Bagel at $8\times$), which is drastically smaller than the $-44.8\%$ gap observed between Reasoning-Guided Generation and De-contextualized Generation.
Moreover, the evaluated UMMs are explicitly trained for reasoning-guided image generation. We further examine the training length distributions of representative models (Appx.~\ref{appx:length_stats}) and verify that the \bench{} evaluation sequences fall well within their typical training ranges, indicating no out-of-distribution length.
We further conduct controlled experiments examining refined-prompt quality, prompt format, delimiters, testmini sampling, and the effect of preceding reasoning content with $R_p$ held fixed. These results show that the examined factors cannot fully explain the gap and that preceding reasoning content still degrades performance when $R_p$ is fixed; full results are provided in Appx.~\ref{appx:additional_controls}.

\begin{table}[t]
\centering
\small
\setlength{\tabcolsep}{1.5pt}
\begin{tabular}{llcccccc}
\toprule
\textbf{Model} & \textbf{\#$R_p$} & \textbf{Code} & \textbf{Arith.} 
& \textbf{Spatial} & \textbf{Attr.} & \textbf{Text} & \textbf{Overall} \\
\midrule
\multirow{3}{*}{Bagel}
  & $1\times$ & 58.0 & 51.0 & 58.0 & 60.0 & 86.0 & 62.6 \\
  & $4\times$ & 63.0 & 49.0 & 52.0 & 53.0 & 85.0 & 60.4 \\
  & $8\times$ & 58.0 & 46.0 & 48.0 & 47.0 & 84.0 & 56.6 \\
\midrule
\multirow{3}{*}{ThinkMorph}
  & $1\times$ & 49.0 & 37.0 & 45.0 & 55.0 & 71.0 & 51.4 \\
  & $4\times$ & 46.0 & 34.0 & 37.0 & 55.0 & 73.0 & 49.0 \\
  & $8\times$ & 44.0 & 33.0 & 34.0 & 51.0 & 73.0 & 47.0 \\
\midrule
\multirow{3}{*}{\makecell[l]{Bagel-Zebra-\\CoT}}
  & $1\times$ & 50.0 & 43.0 & 33.0 & 48.0 & 85.0 & 51.8 \\
  & $4\times$ & 53.0 & 38.0 & 30.0 & 42.0 & 77.0 & 48.0 \\
  & $8\times$ & 49.0 & 36.0 & 25.0 & 38.0 & 74.0 & 44.4 \\
\bottomrule
\end{tabular}
\caption{Length-controlled ablation. $1\times$/$4\times$/$8\times$ 
denotes the refined prompt $R_p$ repeated once, four, or eight 
times, where $4\times$ approximates the average token length 
of a full reasoning trace in our evaluation.}
\label{tab:length_control}
\end{table}

\paragraph{Experiments on Alternative Evaluators.}
\label{sec:alternative_evaluators}
\begin{table*}[th]
\centering
\resizebox{\textwidth}{!}{
\setlength{\tabcolsep}{7pt}
\begin{tabular}{l|c|rrrrrrrrrr|rr}
\toprule
\multirow{2}{*}{\textbf{Evaluator}} & \multirow{2}{*}{\textbf{Setting}} & \multicolumn{2}{|c}{\textbf{Code}} & \multicolumn{2}{c}{\textbf{Arithmetic}} & \multicolumn{2}{c}{\textbf{Spatial}} & \multicolumn{2}{c}{\textbf{Attribute}} & \multicolumn{2}{c}{\textbf{Text}} & \multicolumn{2}{|c}{\textbf{Overall}} \\
\cmidrule(lr){3-4} \cmidrule(lr){5-6} \cmidrule(lr){7-8} \cmidrule(lr){9-10} \cmidrule(lr){11-12} \cmidrule(lr){13-14}
& & \multicolumn{1}{|c}{Acc} & \multicolumn{1}{c}{$\Delta$} & \multicolumn{1}{c}{Acc} & \multicolumn{1}{c}{$\Delta$} & \multicolumn{1}{c}{Acc} & \multicolumn{1}{c}{$\Delta$} & \multicolumn{1}{c}{Acc} & \multicolumn{1}{c}{$\Delta$} & \multicolumn{1}{c}{Acc} & \multicolumn{1}{c}{$\Delta$} & \multicolumn{1}{|c}{Acc} & \multicolumn{1}{c}{$\Delta$} \\ \midrule
\multirow{3}{*}{\makecell[l]{Qwen3-VL\\-235B-A22B}} 
 & \badgeD & 12.0 & \dt{-} & 3.0 & \dt{-} & 6.0 & \dt{-} & 12.0 & \dt{-} & 8.0 & \dt{-} & 8.2 & \dt{-} \\
 & \badgeC & 33.0 & \dt{+21.0} & 18.0 & \dt{+15.0} & 26.0 & \dt{+20.0} & 21.0 & \dt{+9.0} & 12.0 & \dt{+4.0} & 22.0 & \dt{+13.8} \\
 & \badgeA & \textbf{57.0} & \dt{\textbf{+24.0}} & \textbf{42.0} & \dt{\textbf{+24.0}} & \textbf{50.0} & \dt{\textbf{+24.0}} & \textbf{42.0} & \dt{\textbf{+21.0}} & \textbf{52.0} & \dt{\textbf{+40.0}} & \textbf{48.6} & \dt{\textbf{+26.6}} \\ \midrule
\multirow{3}{*}{\makecell[l]{Gemini-2.5\\-Pro}} 
 & \badgeD & 10.0 & \dt{-} & 4.0 & \dt{-} & 7.0 & \dt{-} & 10.0 & \dt{-} & 9.0 & \dt{-} & 8.0 & \dt{-} \\
 & \badgeC & 30.0 & \dt{+20.0} & 18.0 & \dt{+14.0} & 23.0 & \dt{+16.0} & 17.0 & \dt{+7.0} & 9.0 & \dt{0.0} & 19.4 & \dt{+11.4} \\
 & \badgeA & \textbf{53.0} & \dt{\textbf{+23.0}} & \textbf{40.0} & \dt{\textbf{+22.0}} & \textbf{47.0} & \dt{\textbf{+24.0}} & \textbf{43.0} & \dt{\textbf{+26.0}} & \textbf{48.0} & \dt{\textbf{+39.0}} & \textbf{46.2} & \dt{\textbf{+26.8}} \\ \midrule
\multirow{3}{*}{\makecell[l]{Gemini-3.1\\-Pro}} 
 & \badgeD & 10.0 & \dt{-} & 3.0 & \dt{-} & 6.0 & \dt{-} & 10.0 & \dt{-} & 8.0 & \dt{-} & 7.4 & \dt{-} \\
 & \badgeC & 31.0 & \dt{+21.0} & 16.0 & \dt{+13.0} & 23.0 & \dt{+17.0} & 15.0 & \dt{+5.0} & 10.0 & \dt{+2.0} & 19.0 & \dt{+11.6} \\
 & \badgeA & \textbf{52.0} & \dt{\textbf{+21.0}} & \textbf{41.0} & \dt{\textbf{+25.0}} & \textbf{48.0} & \dt{\textbf{+25.0}} & \textbf{40.0} & \dt{\textbf{+25.0}} & \textbf{45.0} & \dt{\textbf{+35.0}} & \textbf{45.2} & \dt{\textbf{+26.2}} \\ \bottomrule
\end{tabular}
}
\caption{Alternative evaluator results on the UniCoT model. Acc and $\Delta$ denote visual verification accuracy (\%) and performance gain over the previous setting, respectively. \badgeD, \badgeC, and \badgeA represent Direct Generation, Reasoning-Guided Generation and De-contextualized Generation, respectively.}
\label{tab:cross_eval}
\end{table*}

To further assess the robustness of our evaluation results, we employ two additional closed-source models, Gemini-2.5-Pro and Gemini-3.1-Pro, as alternative automated evaluators. We use these evaluators to assess UniCoT under all three settings. As shown in Table~\ref{tab:cross_eval}, minor absolute differences exist, possibly reflecting differences in the evaluators' baseline VQA capabilities (human agreement: 0.941 and 0.950 for Gemini-2.5-Pro and Gemini-3.1-Pro, respectively, compared with 0.924 for Qwen3-VL-235B-A22B); nevertheless, the relative performance trends across all three settings remain highly consistent. Gemini-2.5-Pro and Gemini-3.1-Pro replicate the experimental findings, showing substantial performance gains in De-contextualized Generation over Reasoning-Guided Generation. These consistent trends suggest that the observed reasoning-to-generation gap is not an artifact of a particular evaluator.

\paragraph{Gold Refined-Prompt Baseline.}

\begin{table}[t]
\centering
\small
\setlength{\tabcolsep}{3pt}
\begin{tabular}{lcccccc}
\toprule
\textbf{Prompt} & \textbf{Code} & \textbf{Arith.} & \textbf{Spatial} & \textbf{Attr.} & \textbf{Text} & \textbf{Overall} \\
\midrule
Model $R_p$ & 58.0 & 51.0 & 58.0 & 60.0 & 86.0 & 62.6 \\
Gold $R_p^{*}$ & \textbf{72.0} & \textbf{58.0} & \textbf{66.0} & \textbf{61.0} & \textbf{91.0} & \textbf{69.6} \\
\bottomrule
\end{tabular}
\caption{Bagel performance (\%) with its model-generated refined prompt $R_p$ and the human-verified, answer-grounded gold refined prompt $R_p^{*}$.}
\label{tab:gold_prompt}
\end{table}

To separate generation fidelity from errors in a model's self-generated $R_p$, we construct an answer-grounded gold refined prompt $R_p^{*}$ for each instance. Gemini-3.1-Pro receives the task and human-annotated ground-truth criterion and produces a self-contained visual description containing every required element and only the final state. Each description is human-verified against the criterion before use. As shown in Tab.~\ref{tab:gold_prompt}, Bagel improves by 7.0 points over its own $R_p$, from 62.6\% to 69.6\%. This moderate gain is consistent with Bagel's high reasoning-chain accuracy and leaves substantial headroom in the generation stage, where the model must faithfully translate the intended visual semantics specified in the refined prompt into pixel-level realization.

\paragraph{Attention Analysis.}
\label{sec:discussion_attention_analysis}
\begin{table}[t]
\centering
\small
\begin{tabular}{lccc}
\toprule
\textbf{Layers} & \textbf{$P$} & \textbf{$R_t$} & \textbf{$R_p$} \\
\midrule
1--7  & 3.98 & 5.50 & 9.79 \\
8--14  & 3.11 & 4.22 & 8.24 \\
15--21 & 0.20 & 0.30 & 0.79 \\
22--28 & 0.12 & 0.21 & 0.58 \\
\midrule
\textbf{1--28} & \textbf{1.85} & \textbf{2.56} & \textbf{4.85} \\
\bottomrule
\end{tabular}
\caption{Average attention weights across layers for Bagel during image generation, scaled by $10^{-4}$.}
\vspace{-1.0em}
\label{tab:attention}
\end{table}

To better understand why the same intended visual semantics can lead to different generation outcomes, we conduct an attention analysis on Bagel.
During image generation, we extract average attention weights assigned to three token groups: $P$, $R_t$, and $R_p$.
As shown in Table~\ref{tab:attention}, the model allocates substantial attention to $P$ and $R_t$, particularly in early to middle layers.
Across layers, the attention paid to the intermediate reasoning trace $R_t$ remains more than half of that paid to $R_p$.
This suggests that image generation is highly sensitive to contextual interference, which may compete with the final visual specification and weaken the transfer from intended visual semantics to pixels.
This suggests that contextual interference may partly contribute to the observed \alignment{} gap in UMMs.
Appx.~\ref{appx:attention_analysis} provides further details, including analogous analyses on ThinkMorph and T2I-R1, which reveal similar attention patterns and further support the presence of contextual interference across different UMMs.
We hope these findings motivate future work that better preserves intended visual semantics during reasoning-to-generation transfer.

\paragraph{Error Analysis.}
\label{discussion:error_analysis}

To understand failure modes, error cases from Bagel under Setting 2 are analyzed, identifying four error types: (1) \textit{Reasoning Errors} (5.8\%; see Fig.~\ref{fig:error_type1}): the model produces an incorrect reasoning chain or derives an incorrect final state, such as miscalculating object counts; (2) \textit{Instruction Misinterpretation} (10.6\%; see Fig.~\ref{fig:error_type2}): the model misunderstands the fundamental intent of the instruction, such as rendering code text itself rather than visualizing the object specified by the code; (3) \textit{Concept Hallucination} (8.2\%; see Fig.~\ref{fig:error_type3}): the generated image introduces objects or visual concepts that are not specified in the input prompt; and (4) \textit{Task-Specific Errors} (75.4\%; see Figs.~\ref{fig:error_type42}--\ref{fig:error_type44}): the model correctly reasons about the intended visual outcome but fails to faithfully realize specific task requirements in the generated image, such as object counts, spatial arrangements, attributes, or rendered text. The dominance of task-specific errors highlights the \alignment{} gap: even when models derive the correct visual intent in textual reasoning, they often fail to faithfully realize it in the generated image. This result demonstrates \bench{}'s effectiveness as a diagnostic benchmark for fine-grained failure analysis. Representative error cases are provided in Appx.~\ref{appx:error_cases}.

\section{Related Work}

\textbf{Unified Multimodal Models.}
Unified multimodal models aim to support multimodal understanding and generation within a single model, typically by mapping text and images into a shared representational interface.
Recent approaches span diffusion-based models~\cite{li2025dual,swerdlow2025unified,shi2025muddit}, autoregressive models~\cite{team2024chameleon,wang2024emu3,wu2025qwen,chen2025janus,tong2025metamorph}, and hybrids that combine both mechanisms~\cite{zhou2024transfusion,xie2024show,deng2025emerging}.
Despite these advances, characterizing cross-modal interactions and the interplay between understanding and generation remains an active research area~\cite{yan2025can,liang2025rover,niu2025does,zhang2025generative}.
\bench{} complements this line of work by providing a diagnostic evaluation of \alignment{}, specifically how textual reasoning influences visual generation.

\paragraph{Chain-of-Thought.} 
Chain-of-thought reasoning has emerged as a powerful technique to enhance the capabilities of LLMs~\cite{wei2022chain,chen2025towards} and MLLMs~\cite{li2025perception,luo2025unlocking}. Recent large reasoning models~\citep{openai2024learning,openai2025o3mini,guo2025deepseek} further demonstrate that test-time scaling enables more accurate outcomes.
UMMs integrate processing for language and image within a single architecture, which provides a natural foundation for reasoning-guided image generation. Consequently, recent works~\citep{jiang2025t2i,deng2025emerging,jin2025srum,yang2025chartmimic,qin2025uni,liang2025rover,gao2026text} have adopted  explicit reasoning chains to plan via natural language before synthesizing images. Despite the appeal of this reasoning-guided paradigm, the actual alignment between reasoning and visual generation quality remains underexplored, motivating our investigation of reasoning-to-generation alignment in this area.

\paragraph{T2I Benchmarks.}
T2I benchmarks have progressed from evaluating explicit prompt adherence to probing implicit reasoning capabilities.
Prior work focuses on generation quality via text-image alignment~\cite{saharia2022photorealistic,ghosh2023geneval,lee2023holistic}, compositional generation~\cite{huang2023t2i}, and safety constraints~\cite{schramowski2023safe,seshadri2024bias}.
More recent benchmarks target reasoning-driven scenarios that require commonsense and world knowledge~\cite{fu2024commonsense,niu2025wise,chen2025r2i,sun2025t2i,chang2025oneigbench}.
\bench{} complements this line of work with probing cross-modality alignment between textual reasoning and visual generation in UMMs.

\section{Conclusion}
We introduce \bench{}, a diagnostic benchmark for reasoning-guided image generation in UMMs, with $5$ tasks, $2,000$ instances and a controlled framework comparing direct, reasoning-guided, and de-contextualized generation.
Across $8$ open-source models, we observe that reasoning improves over direct prompting, the intended visual semantics expressed in textual reasoning are not always faithfully reflected in the generated images, and conditioning only on the refined prompt often performs best, showing the \alignment{} gap.
These findings suggest that advancing UMMs requires not only stronger reasoning capabilities, but also more robust mechanisms to ensure that inferred visual semantics can be reliably carried through the image generation process.

\section*{Limitations}
\bench{} focuses on single-turn reasoning-guided image generation, which allows us to cleanly attribute each outcome to a specific reasoning-to-generation (i.e., from textual reasoning to visual generation) process. We do not explicitly evaluate image editing tasks. Nevertheless, we expect similar reasoning-to-generation tensions to arise in editing pipelines that condition on verbose reasoning or edit histories. Extending \bench{} and our diagnostic protocol to multi-turn interactions remains an important direction, as interference effects may become more pronounced when reasoning traces accumulate over turns.

\section*{Acknowledgements}
This research was supported in part by the NSF under grants 2146151 and 2422214.
We thank the reviewers, Area Chairs, Senior Area Chairs, and Program Chairs for their constructive feedback and suggestions during the review process.

\bibliography{custom}

\etocdepthtag.toc{appendixtag}
\clearpage
\appendix

\etocsettagdepth{mainmatter}{none}
\etocsettagdepth{appendixtag}{subsection}
\etocsettocstyle{\section*{Appendix}}{}
\tableofcontents

\clearpage

\section{LLM Disclosure}
We use Gemini-3-Pro to conduct LLM-Assisted Data Augmentation as detailed in 
Section~\ref{app:data_curation}. LLMs are used to assist with drafting 
and polishing paper text. No LLMs are used to originate research ideas.

\section{Data Annotation}
\label{app:data_curation}
\subsection{Task Taxonomy}
\begin{figure}[t]
    \centering
    \includegraphics[width=0.5\textwidth]{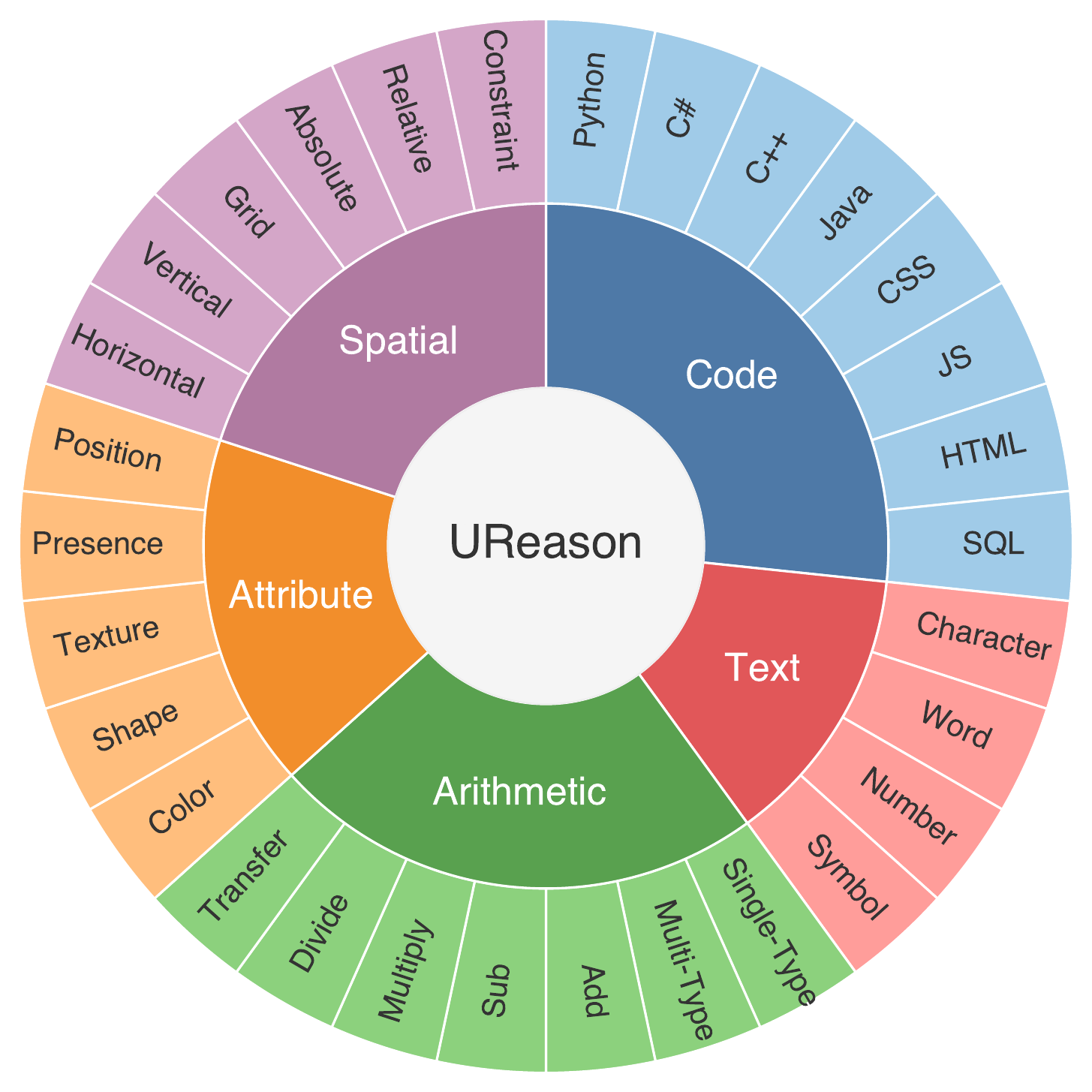}
    \caption{Taxonomy of \bench{} tasks. The benchmark contains $5$ task categories with $30$ fine-grained subcategories covering diverse reasoning and visual generation challenges.}
    \label{fig:subcategory}
\end{figure}

To guarantee the quality and diversity of \bench{}, human experts first establish a fine-grained taxonomy spanning 30 subcategories under the five primary tasks (Fig.~\ref{fig:subcategory}). Each subcategory is designed to isolate a distinct reasoning capability or visual aspect, intentionally focusing on fundamental operations and visual characteristics. This granular design enables convenient and precise error localization and diagnostic analysis. Importantly, these subcategories are not mutually exclusive and can be combined to construct test cases when needed.

\paragraph{\textsc{Code Reasoning}}

\bench{} defines eight subcategories for Code Reasoning based on programming languages, covering object-oriented languages (Python, C\#, C++, Java), front-end technologies (HTML, CSS, JavaScript), SQL for data querying, and various programming paradigms. Models are required to understand language-specific syntax, control flow, data structures, and computational logic to simulate code execution and determine the final visual rendering. This comprehensive coverage ensures that models must possess broad code comprehension capabilities across different programming ecosystems. Notably, code snippets are designed such that their visual outputs involve arithmetic counts, spatial layouts, attribute constraints, and text rendering.

\paragraph{\textsc{Arithmetic Reasoning}}

\bench{} defines seven subcategories for Arithmetic Reasoning covering different operational complexities and object configurations. Single-Type~(St) and Multi-Type~(Mt) subcategories distinguish scenarios based on object diversity, while Add~(Add), Subtract~(Sub), Multiply~(Mul), Divide~(Div), and Transfer~(Trf) subcategories focus on specific arithmetic operations. Models are required to perform sequential numerical reasoning through narrative events, tracking quantity changes dynamically and ensuring that the final visual output strictly reflects the calculated object count. This demands models to transcend superficial keyword matching and function as quantitative reasoners.

\paragraph{\textsc{Spatial Reasoning}}

\bench{} defines six subcategories for Spatial Reasoning covering different spatial arrangement paradigms: Horizontal~(Hor), Vertical~(Ver), Grid~(Grd), Absolute~(Abs), Relative~(Rel), and Constraint~(Cst). These subcategories assess models' capacity to resolve high-level semantic descriptions into structured coordinate-based layouts. Unlike standard benchmarks where spatial relationships are explicitly stated, models must infer spatial configurations from implicit cues, logical constraints, and relational reasoning. The Constraint subcategory particularly challenges models to act as spatial reasoners that interpret and satisfy predefined placement rules and restrictions—such as ``object A cannot be adjacent to object B'' or ``all red objects must be on the left side''—before determining the final spatial arrangement that complies with all specified constraints.

\paragraph{\textsc{Attribute Reasoning}}

\bench{} defines five subcategories for Attribute Reasoning based on different object properties: Color~(Clr), Shape~(Shp), Texture~(Tex), Presence~(Prs) and Position~(Pos). These subcategories evaluate models' capability to track and update object attributes through state transitions and logical modifications described in the text. Models must perform logical filtering to derive the terminal outcome rather than rendering initial or intermediate states, effectively suppressing visual biases toward irrelevant attributes mentioned in the prompt. This requires maintaining attribute consistency throughout complex state evolution narratives.

\paragraph{\textsc{Text Reasoning}}

\bench{} defines four subcategories for Text Reasoning based on the granularity of textual elements: Character~(Chr), Word~(Wrd), Number~(Num), and Symbol~(Sym). These subcategories assess models' ability to perform context-aware text inference at different semantic levels. Models must derive the target text through contextual rules, linguistic transformations, mathematical operations, or symbolic reasoning, rather than directly rendering explicitly quoted strings. This requires models to act as symbolic reasoners that suppress irrelevant information and render only the logically derived final result.

\subsection{Annotation Pipeline}

In this section, we provide comprehensive details about the data curation process of \bench{}, including the human-curated seed data construction, LLM-assisted augmentation pipeline, and data statistics.

\subsubsection{Human-Curated Seed Data Construction}

\paragraph{Annotator Background.}
Our annotation team consists of 5 expert annotators with professional backgrounds in computer vision and natural language processing. All annotators possess at least a Master's degree in computer science, with an average of 3 years of experience in AI research. 

\paragraph{Taxonomy Design Process.}
The design of our fine-grained taxonomy follows a systematic, multi-stage approach. We first identify five primary reasoning dimensions—Code, Arithmetic, Spatial, Attribute, and Text—based on a literature review of reasoning capabilities in language models and an analysis of real-world visual generation requirements. For each primary category, we conduct brainstorming sessions with human annotators to identify representative subcategories, with selection criteria emphasizing coverage of diverse reasoning patterns, distinctiveness in targeting specific skills, verifiability for objective evaluation and practical relevance to real-world use cases. We then conduct a pilot study with 10 instances per subcategory to verify feasibility. After iterative refinement, we establish the final taxonomy comprising $30$ subcategories across $5$ main tasks, as illustrated in Fig.~\ref{fig:subcategory}.

\paragraph{Seed Instance Construction Details.}
Expert annotators manually construct seed instances following specific design principles:

As shown in Fig.~\ref{fig:task_example}, for prompt design, each prompt necessitates intermediate reasoning steps to derive the visual target. While the target remains implicit, the prompt provides sufficient information for a unique, deterministic solution. For evaluation criterion design, each criterion specifies a concrete, verifiable aspect of the generated image, such as exact object counts, specific text strings, or precise spatial relationships. We specify exact values rather than ranges for quantitative attributes and define clear verification rules for qualitative attributes. Different task categories employ tailored criterion formats: Arithmetic tasks specify exact object counts after all operations, Spatial tasks define precise positional relationships or arrangements, Attribute tasks indicate final resolved object attributes, Text tasks require exact text strings to be rendered, and Code tasks encompass all four of the above aspects.

Notably, our early pilot experiments reveal that excessively large quantities or lengthy text strings pose significant challenges for current UMMs, prompting us to adjust the difficulty levels accordingly in our annotation process.

\subsubsection{LLM-Assisted Data Augmentation}

We use Gemini-3-Pro~\cite{deepmind_gemini3_2025} to generate $10$ candidate variants for each of the $500$ seed instances, yielding $5{,}000$ candidates. We systematically vary three dimensions while preserving a unique answer: reasoning complexity, target visual entities, and narrative context. Complexity is controlled by adding or removing reasoning steps and distractors; visual variation includes object types, quantities, colors, shapes, sizes, and textures; narrative variation covers linguistic styles and everyday or scientific scenarios.
Gemini is used only to propose candidates; all inclusion decisions and final labels are made by human annotators. We also constrain quantities and text lengths based on pilot studies so that the resulting criteria remain visually verifiable.

\paragraph{Human-LLM Interaction Workflow.}
Each of the $5{,}000$ candidates is independently labeled by three expert annotators as accepted, requiring revision, or rejected. Inter-annotator agreement is substantial (Fleiss' $\kappa=0.78$). Disagreements are resolved through discussion, and annotators retain the best three candidates per seed, corresponding to a 30\% adoption rate; the remaining 70\% are not adopted.

Candidates requiring revision enter a multi-round human--LLM refinement loop. Every retained candidate then undergoes unanimous final human verification for logical correctness, label correctness, diversity, stylistic variation, and compliance with the evaluation criterion. Together with the $500$ manually authored seeds, all $1{,}500$ retained variants pass this final verification to form the $2{,}000$-instance benchmark. We additionally inspect candidates for shortcut cues and superficial patterns that could reveal the target without the intended reasoning.

\subsection{Data Statistics}
\begin{table*}[h]
    \centering
    \small 
    \begin{tabular}{l|ccccc|c}
        \toprule
        Category & Code & Arithmetic & Spatial & Attribute & Text & Overall \\
        \midrule
        Count              & 400   & 400 & 400 & 400 & 400 & 2000 \\
        Subcategories      & 8    & 7 & 6 & 5 & 4 & 30 \\
        Prompt Length (AVG.)      & 113.14 & 131.44 & 169.27 & 97.78 & 79.66 & 118.26 \\
        Prompt Length (STD.)      & 51.61 & 13.11 & 35.90 & 18.79 & 8.33 & 42.97 \\
        \bottomrule
    \end{tabular}
    \caption{Statistics of \bench{} across different tasks.}
    \label{tab:data_stat}
\end{table*}
Tab.~\ref{tab:data_stat} presents the statistics of \bench{}. The benchmark comprises $2,000$ instances evenly distributed across five task categories ($400$ each), spanning $30$ fine-grained subcategories. Subcategory counts reflect each domain's scope: \textsc{Code} encompasses $8$ programming languages, while \textsc{Arithmetic}, \textsc{Spatial}, \textsc{Attribute}, and \textsc{Text} contain $7$, $6$, $5$, and $4$ subcategories respectively. Prompt lengths are measured using the Qwen3-8B tokenizer.

\subsection{Details on Comparison with Existing Benchmarks}
\label{appx:benchmark_comparison}

Tab.~\ref{tab:comparison} compares \bench{} with existing reasoning-driven image generation benchmarks. Compared with prior text-to-image and image-editing benchmarks, \bench{} provides a larger and more fine-grained diagnostic testbed, containing 2,000 instances across 5 task categories and 30 subcategories.
This scale and taxonomy allow \bench{} to cover a broader range of reasoning-to-generation challenges, including \textsc{Code}, \textsc{Arithmetic}, \textsc{Spatial}, \textsc{Attribute}, and \textsc{Text} reasoning.
In particular, \bench{} introduces the under-explored \textsc{Code} reasoning domain, where models must interpret code snippets across multiple programming languages, reason over execution logic or state changes, and synthesize the corresponding visual outcomes.

Beyond dataset coverage, \bench{} differs from prior benchmarks in how it measures reasoning-to-generation alignment.
Most existing benchmarks primarily evaluate final-image performance under direct generation, and some only run preliminary reasoning-chain experiments on specific models.
In contrast, \bench{} evaluates whether the image generated in the same inference process faithfully reflects the model's own textual reasoning.
By further comparing Reasoning-Guided Generation with De-contextualized Generation, \bench{} provides a controlled way to measure whether intermediate reasoning context helps or interferes with visual synthesis. 
This formulation differs from benchmarks that infer alignment or consistency indirectly from final task performance across separate multi-modality understanding and generation tasks~\citep{li2026girbench}.
Therefore, \bench{} contributes both a broad benchmark with new task coverage and a controlled protocol for analyzing reasoning-to-generation alignment in UMMs.

\section{Details on Evaluation Settings}
\label{app:details_eval_settings}

Our evaluation settings are designed to diagnose whether visual generation in reasoning-guided image generation faithfully reflects the intended visual semantics expressed in textual reasoning, thereby providing a controlled lens on \alignment{} in UMMs.

UMMs typically support two generation modes: \emph{Direct Generation}, which generates an image directly from the user instruction, and \emph{Reasoning-Guided Generation}, which generates textual reasoning for image generation. We additionally introduce \emph{De-contextualized Generation}, which conditions image generation only on the model's intended visual specification\footnote{In this paper, we use \emph{intended visual semantics} to refer to the final visual target in the abstract representational space, and \emph{intended visual specification} to refer to its expression in concrete textual form.}.

A key design choice is how to obtain this intended visual specification.
Rather than relying on an external model or human annotation to summarize the reasoning trace, we explicitly require the evaluated model to output a refined prompt, $R_p$, at the end of the reasoning trace.
This refined prompt serves as the model's own textual summary of the final visual target, i.e., the intended visual semantics that the model aims to realize in the image.

This design is important for fairness and comparability.
In principle, one could attempt to infer the intended visual semantics directly from the textual reasoning trace.
However, doing so would typically require an additional summarization step, often involving an external model, to convert reasoning trace into a concise visual specification.
Such a multi-stage pipeline would introduce additional model bias and cumulative error, making it harder to attribute performance differences to the evaluated UMM itself.
By instead requiring the model to explicitly produce $R_p$, we ensure that the de-contextualized setting uses the model's own final visual specification while avoiding confounding effects from external summarization.

This formulation also brings practical benefits.
Because $R_p$ is explicitly expressed in text, it provides an interpretable representation of the intended visual semantics for downstream analysis.
This supports not only a fair comparison between Reasoning-Guided Generation and De-contextualized Generation, but also subsequent analyses of reasoning chains and attention behavior discussed in the paper.

\section{\texorpdfstring{Discussion on the Scope of Reasoning}{Discussion on the Scope of Reasoning}}
\label{appx:reasoning_scope}

In this work, reasoning refers to the free-form, linear solution traces produced by existing reasoning-guided UMMs before image synthesis. Self-generated reflection also does not necessarily induce renewed visual grounding~\cite{shi2026vlms}. We do not study structured reasoning frameworks that rely on explicit search, path selection, or process-level verification, such as Tree of Thoughts or Monte Carlo tree search. Existing UMMs are trained primarily on free-form reasoning data, so imposing a structured format only at inference time could introduce a train--test mismatch. Extending unified models with structured and verifiable reasoning remains an important direction for future work.

\section{\texorpdfstring{Additional Controlled Analyses}{Additional Controlled Analyses}}
\label{appx:additional_controls}

\paragraph{Prompt-Format and Delimiter Controls.}
We test whether the Setting-2 gap is an artifact of the output template by changing only its format to an equation-style representation, \texttt{thinking="...", refined\_prompt="..."}. We separately prepend empty \texttt{<think></think>} delimiters to the same $R_p$ in Setting 3. Seeds, samplers, and all sampling parameters are fixed within each comparison. Tab.~\ref{tab:format_delimiter} shows that the format changes overall accuracy by at most 0.8 points and the delimiters by at most 2.4 points, both far smaller than the Setting-2-to-Setting-3 gaps.

\paragraph{Holding the Refined Prompt Fixed.}
Starting from Bagel's Setting-2 output, we parse the trace into $[R_t,R_p]$, retain the same $R_p$ and generation format, and replace $R_t$ with a trace sampled from an unrelated instance generated by the same model. Tab.~\ref{tab:trace_intervention} compares $R_p$ alone, the original $R_t+R_p$, and the unrelated $R_t+R_p$. Adding the original trace reduces accuracy from 62.6\% to 21.0\%; an unrelated trace reduces it further to 16.4\%. This intervention isolates the effect of preceding reasoning content while controlling for the refined prompt, delimiters, format, and generation mode.

\paragraph{Subsampling Stability.}
We randomly draw 100 instances per task from the full 2,000-instance benchmark and recompute overall accuracy on each resulting 500-instance subset, repeating the procedure 2,000 times. For Bagel, the subsampled accuracies are $64.1\pm2.3$\% under Setting 3 and $19.5\pm2.1$\% under Setting 2. This variability is much smaller than the observed 44.8-point gap, showing that the result is stable to the particular testmini sample.

\begin{table*}[t]
\centering
\small
\setlength{\tabcolsep}{5pt}
\begin{tabular}{llcccccc}
\toprule
\textbf{Control} & \textbf{Model/Condition} & \textbf{Code} & \textbf{Arith.} & \textbf{Spatial} & \textbf{Attr.} & \textbf{Text} & \textbf{Overall} \\
\midrule
\multirow{4}{*}{Prompt format}
 & Bagel, original & 30.0 & 14.0 & 15.0 & 19.0 & 11.0 & 17.8 \\
 & Bagel, equation format & 29.0 & 17.0 & 13.0 & 16.0 & 12.0 & 17.4 \\
 & UniCoT-v2, original & 27.0 & 19.0 & 29.0 & 21.0 & 21.0 & 23.4 \\
 & UniCoT-v2, equation format & 25.0 & 19.0 & 27.0 & 20.0 & 22.0 & 22.6 \\
\midrule
\multirow{4}{*}{Delimiter}
 & Bagel, $R_p$ only & 58.0 & 51.0 & 58.0 & 60.0 & 86.0 & 62.6 \\
 & Bagel, empty delimiters $+R_p$ & 54.0 & 49.0 & 58.0 & 61.0 & 83.0 & 61.0 \\
 & UniCoT-v2, $R_p$ only & 65.0 & 45.0 & 46.0 & 67.0 & 87.0 & 62.0 \\
 & UniCoT-v2, empty delimiters $+R_p$ & 63.0 & 43.0 & 43.0 & 64.0 & 85.0 & 59.6 \\
\bottomrule
\end{tabular}
\caption{Prompt-format and delimiter controls on testmini. The equation format uses \texttt{thinking="...", refined\_prompt="..."}; the delimiter control prepends \texttt{<think></think>} to the same $R_p$. Seeds, samplers, and sampling parameters are held fixed within each comparison.}
\label{tab:format_delimiter}
\end{table*}

\begin{table}[t]
\centering
\small
\setlength{\tabcolsep}{3pt}
\resizebox{\columnwidth}{!}{%
\begin{tabular}{lcccccc}
\toprule
\textbf{Context} & \textbf{Code} & \textbf{Arith.} & \textbf{Spatial} & \textbf{Attr.} & \textbf{Text} & \textbf{Overall} \\
\midrule
$R_p$ only & 58.0 & 51.0 & 58.0 & 60.0 & 86.0 & 62.6 \\
Original $R_t+R_p$ & 35.0 & 15.0 & 20.0 & 21.0 & 14.0 & 21.0 \\
Unrelated $R_t+R_p$ & 31.0 & 13.0 & 15.0 & 15.0 & 8.0 & 16.4 \\
\bottomrule
\end{tabular}
}
\caption{Controlled intervention on Bagel (accuracy, \%). The refined prompt and generation format are fixed while the preceding reasoning content is varied.}
\label{tab:trace_intervention}
\end{table}

\section{\texorpdfstring{Consistency Between the Test and Testmini Sets}{Consistency Between the Test and Testmini Sets}}
\label{appx:textmini}
\begin{table*}[t]
\centering
\resizebox{\textwidth}{!}{
\setlength{\tabcolsep}{5pt}
\begin{tabular}{l|l|c|rrrrrrrrrr|rr}
\toprule

\multirow{2}{*}{\textbf{Model}} & \multirow{2}{*}{\textbf{Test Set}} & \multirow{2}{*}{\textbf{Setting}} & \multicolumn{2}{|c}{\textbf{Code}} & \multicolumn{2}{c}{\textbf{Arithmetic}} & \multicolumn{2}{c}{\textbf{Spatial}} & \multicolumn{2}{c}{\textbf{Attribute}} & \multicolumn{2}{c}{\textbf{Text}} & \multicolumn{2}{|c}{\textbf{Overall}} \\
\cmidrule(lr){4-5} \cmidrule(lr){6-7} \cmidrule(lr){8-9} \cmidrule(lr){10-11} \cmidrule(lr){12-13} \cmidrule(lr){14-15}

& & & \multicolumn{1}{|c}{Acc} & \multicolumn{1}{c}{$\Delta$} & \multicolumn{1}{c}{Acc} & \multicolumn{1}{c}{$\Delta$} & \multicolumn{1}{c}{Acc} & \multicolumn{1}{c}{$\Delta$} & \multicolumn{1}{c}{Acc} & \multicolumn{1}{c}{$\Delta$} & \multicolumn{1}{c}{Acc} & \multicolumn{1}{c}{$\Delta$} & \multicolumn{1}{|c}{Acc} & \multicolumn{1}{c}{$\Delta$} \\ \midrule

\multirow{6}{*}{Bagel} & \multirow{3}{*}{testmini} & \badgeD & 10.0 & \dt{-} & 5.0 & \dt{-} & 3.0 & \dt{-} & 6.0 & \dt{-} & 9.0 & \dt{-} & 6.6 & \dt{-} \\
 &  & \badgeC & 30.0 & \dt{+20.0} & 14.0 & \dt{+9.0} & 15.0 & \dt{+12.0} & 19.0 & \dt{+13.0} & 11.0 & \dt{+2.0} & 17.8 & \dt{+11.2} \\
 &  & \badgeA & \textbf{58.0} & \dt{\textbf{+28.0}} & \textbf{51.0} & \dt{\textbf{+37.0}} & \textbf{58.0} & \dt{\textbf{+43.0}} & \textbf{60.0} & \dt{\textbf{+41.0}} & \textbf{86.0} & \dt{\textbf{+75.0}} & \textbf{62.6} & \dt{\textbf{+44.8}} \\
\cline{2-15}
 & \multirow{3}{*}{test} & \badgeD & 12.0 & \dt{-} & 8.0 & \dt{-} & 6.0 & \dt{-} & 5.0 & \dt{-} & 10.0 & \dt{-} & 8.2 & \dt{-} \\
 &  & \badgeC & 33.0 & \dt{+21.0} & 18.0 & \dt{+10.0} & 16.0 & \dt{+10.0} & 21.0 & \dt{+16.0} & 13.0 & \dt{+3.0} & 20.2 & \dt{+12.0} \\
 &  & \badgeA & \textbf{60.0} & \dt{\textbf{+27.0}} & \textbf{54.0} & \dt{\textbf{+36.0}} & \textbf{59.0} & \dt{\textbf{+43.0}} & \textbf{63.0} & \dt{\textbf{+42.0}} & \textbf{87.0} & \dt{\textbf{+74.0}} & \textbf{64.6} & \dt{\textbf{+44.4}} \\
 \midrule

\multirow{6}{*}{UniCoT} & \multirow{3}{*}{testmini} & \badgeD & 12.0 & \dt{-} & 3.0 & \dt{-} & 6.0 & \dt{-} & 12.0 & \dt{-} & 8.0 & \dt{-} & 8.2 & \dt{-} \\
 &  & \badgeC & 33.0 & \dt{+21.0} & 18.0 & \dt{+15.0} & 26.0 & \dt{+20.0} & 21.0 & \dt{+9.0} & 12.0 & \dt{+4.0} & 22.0 & \dt{+13.8} \\
 &  & \badgeA & \textbf{57.0} & \dt{\textbf{+24.0}} & \textbf{42.0} & \dt{\textbf{+24.0}} & \textbf{50.0} & \dt{\textbf{+24.0}} & \textbf{42.0} & \dt{\textbf{+21.0}} & \textbf{52.0} & \dt{\textbf{+40.0}} & \textbf{48.6} & \dt{\textbf{+26.6}} \\
\cline{2-15}
 & \multirow{3}{*}{test} & \badgeD & 13.0 & \dt{-} & 5.0 & \dt{-} & 9.0 & \dt{-} & 13.0 & \dt{-} & 9.0 & \dt{-} & 9.8 & \dt{-} \\
 &  & \badgeC & 37.0 & \dt{+24.0} & 18.0 & \dt{+13.0} & 31.0 & \dt{+22.0} & 24.0 & \dt{+11.0} & 15.0 & \dt{+6.0} & 25.0 & \dt{+15.2} \\
 &  & \badgeA & \textbf{57.0} & \dt{\textbf{+20.0}} & \textbf{46.0} & \dt{\textbf{+28.0}} & \textbf{54.0} & \dt{\textbf{+23.0}} & \textbf{42.0} & \dt{\textbf{+18.0}} & \textbf{57.0} & \dt{\textbf{+42.0}} & \textbf{51.2} & \dt{\textbf{+26.2}} \\
 \midrule

\multirow{6}{*}{SRUM} & \multirow{3}{*}{testmini} & \badgeD & 11.0 & \dt{-} & 4.0 & \dt{-} & 3.0 & \dt{-} & 6.0 & \dt{-} & 7.0 & \dt{-} & 6.2 & \dt{-} \\
 &  & \badgeC & 25.0 & \dt{+14.0} & 8.0 & \dt{+4.0} & 20.0 & \dt{+17.0} & 24.0 & \dt{+18.0} & 6.0 & \dt{-1.0} & 16.6 & \dt{+10.4} \\
 &  & \badgeA & \textbf{67.0} & \dt{\textbf{+42.0}} & \textbf{50.0} & \dt{\textbf{+42.0}} & \textbf{50.0} & \dt{\textbf{+30.0}} & \textbf{50.0} & \dt{\textbf{+26.0}} & \textbf{82.0} & \dt{\textbf{+76.0}} & \textbf{59.8} & \dt{\textbf{+43.2}} \\
\cline{2-15}
 & \multirow{3}{*}{test} & \badgeD & 10.0 & \dt{-} & 5.0 & \dt{-} & 6.0 & \dt{-} & 10.0 & \dt{-} & 9.0 & \dt{-} & 8.0 & \dt{-} \\
 &  & \badgeC & 27.0 & \dt{+17.0} & 10.0 & \dt{+5.0} & 23.0 & \dt{+17.0} & 28.0 & \dt{+18.0} & 10.0 & \dt{+1.0} & 19.6 & \dt{+11.6} \\
 &  & \badgeA & \textbf{69.0} & \dt{\textbf{+42.0}} & \textbf{49.0} & \dt{\textbf{+39.0}} & \textbf{53.0} & \dt{\textbf{+30.0}} & \textbf{55.0} & \dt{\textbf{+27.0}} & \textbf{85.0} & \dt{\textbf{+75.0}} & \textbf{62.2} & \dt{\textbf{+42.6}} \\
 \midrule

\multirow{6}{*}{ThinkMorph} & \multirow{3}{*}{testmini} & \badgeD & 9.0 & \dt{-} & 1.0 & \dt{-} & 4.0 & \dt{-} & 3.0 & \dt{-} & 10.0 & \dt{-} & 5.4 & \dt{-} \\
 &  & \badgeC & 19.0 & \dt{+10.0} & 12.0 & \dt{+11.0} & 15.0 & \dt{+11.0} & 26.0 & \dt{+23.0} & 5.0 & \dt{-5.0} & 15.4 & \dt{+10.0} \\
 &  & \badgeA & \textbf{49.0} & \dt{\textbf{+30.0}} & \textbf{37.0} & \dt{\textbf{+25.0}} & \textbf{45.0} & \dt{\textbf{+30.0}} & \textbf{55.0} & \dt{\textbf{+29.0}} & \textbf{71.0} & \dt{\textbf{+66.0}} & \textbf{51.4} & \dt{\textbf{+36.0}} \\
\cline{2-15}
 & \multirow{3}{*}{test} & \badgeD & 12.0 & \dt{-} & 3.0 & \dt{-} & 5.0 & \dt{-} & 5.0 & \dt{-} & 13.0 & \dt{-} & 7.6 & \dt{-} \\
 &  & \badgeC & 23.0 & \dt{+11.0} & 16.0 & \dt{+13.0} & 18.0 & \dt{+13.0} & 27.0 & \dt{+22.0} & 9.0 & \dt{-4.0} & 18.6 & \dt{+11.0} \\
 &  & \badgeA & \textbf{48.0} & \dt{\textbf{+25.0}} & \textbf{35.0} & \dt{\textbf{+19.0}} & \textbf{46.0} & \dt{\textbf{+28.0}} & \textbf{53.0} & \dt{\textbf{+26.0}} & \textbf{68.0} & \dt{\textbf{+59.0}} & \textbf{50.0} & \dt{\textbf{+31.4}} \\
\midrule

\multirow{6}{*}{UniCoT-v2} & \multirow{3}{*}{testmini} & \badgeD & 9.0 & \dt{-} & 2.0 & \dt{-} & 2.0 & \dt{-} & 5.0 & \dt{-} & 3.0 & \dt{-} & 4.2 & \dt{-} \\
 & & \badgeC & 27.0 & \dt{+18.0} & 19.0 & \dt{+17.0} & 29.0 & \dt{+27.0} & 21.0 & \dt{+16.0} & 21.0 & \dt{+18.0} & 23.4 & \dt{+19.2} \\
 & & \badgeA & \textbf{65.0} & \dt{\textbf{+38.0}} & \textbf{45.0} & \dt{\textbf{+26.0}} & \textbf{46.0} & \dt{\textbf{+17.0}} & \textbf{67.0} & \dt{\textbf{+46.0}} & \textbf{87.0} & \dt{\textbf{+66.0}} & \textbf{62.0} & \dt{\textbf{+38.6}} \\
\cline{2-15}
 & \multirow{3}{*}{test} & \badgeD & 10.0 & \dt{-} & 4.0 & \dt{-} & 3.0 & \dt{-} & 6.0 & \dt{-} & 4.0 & \dt{-} & 5.4 & \dt{-} \\
 & & \badgeC & 29.0 & \dt{+19.0} & 20.0 & \dt{+16.0} & 31.0 & \dt{+28.0} & 23.0 & \dt{+17.0} & 24.0 & \dt{+20.0} & 25.4 & \dt{+20.0} \\
 & & \badgeA & \textbf{65.0} & \dt{\textbf{+36.0}} & \textbf{48.0} & \dt{\textbf{+28.0}} & \textbf{41.0} & \dt{\textbf{+10.0}} & \textbf{68.0} & \dt{\textbf{+45.0}} & \textbf{84.0} & \dt{\textbf{+60.0}} & \textbf{61.2} & \dt{\textbf{+35.8}} \\
\midrule

\multirow{6}{*}{\makecell[l]{Bagel-Zebra\\-CoT}} & \multirow{3}{*}{testmini} & \badgeD & 7.0 & \dt{-} & 7.0 & \dt{-} & 2.0 & \dt{-} & 10.0 & \dt{-} & 5.0 & \dt{-} & 6.2 & \dt{-} \\
 & & \badgeC & 14.0 & \dt{+7.0} & 10.0 & \dt{+3.0} & 15.0 & \dt{+13.0} & 23.0 & \dt{+13.0} & 10.0 & \dt{+5.0} & 14.4 & \dt{+8.2} \\
 & & \badgeA & \textbf{50.0} & \dt{\textbf{+36.0}} & \textbf{43.0} & \dt{\textbf{+33.0}} & \textbf{33.0} & \dt{\textbf{+18.0}} & \textbf{48.0} & \dt{\textbf{+25.0}} & \textbf{85.0} & \dt{\textbf{+75.0}} & \textbf{51.8} & \dt{\textbf{+37.4}} \\
\cline{2-15}
 & \multirow{3}{*}{test} & \badgeD & 8.0 & \dt{-} & 8.0 & \dt{-} & 3.0 & \dt{-} & 11.0 & \dt{-} & 6.0 & \dt{-} & 7.2 & \dt{-} \\
 & & \badgeC & 16.0 & \dt{+8.0} & 11.0 & \dt{+3.0} & 17.0 & \dt{+14.0} & 26.0 & \dt{+15.0} & 13.0 & \dt{+7.0} & 16.6 & \dt{+9.4} \\
 & & \badgeA & \textbf{52.0} & \dt{\textbf{+36.0}} & \textbf{45.0} & \dt{\textbf{+34.0}} & \textbf{35.0} & \dt{\textbf{+18.0}} & \textbf{50.0} & \dt{\textbf{+24.0}} & \textbf{86.0} & \dt{\textbf{+73.0}} & \textbf{53.6} & \dt{\textbf{+37.0}} \\
\midrule

\multirow{6}{*}{T2I-R1} & \multirow{3}{*}{testmini} & \badgeD & 3.0 & \dt{-} & 6.0 & \dt{-} & 4.0 & \dt{-} & 9.0 & \dt{-} & 2.0 & \dt{-} & 4.8 & \dt{-} \\
 & & \badgeC & 6.0 & \dt{+3.0} & 4.0 & \dt{-2.0} & 2.0 & \dt{-2.0} & 11.0 & \dt{+2.0} & 3.0 & \dt{+1.0} & 5.2 & \dt{+0.4} \\
 & & \badgeA & \textbf{20.0} & \dt{\textbf{+14.0}} & \textbf{15.0} & \dt{\textbf{+11.0}} & \textbf{12.0} & \dt{\textbf{+10.0}} & \textbf{27.0} & \dt{\textbf{+16.0}} & \textbf{47.0} & \dt{\textbf{+44.0}} & \textbf{24.2} & \dt{\textbf{+19.0}} \\
\cline{2-15}
 & \multirow{3}{*}{test} & \badgeD & 4.0 & \dt{-} & 6.0 & \dt{-} & 5.0 & \dt{-} & 10.0 & \dt{-} & 3.0 & \dt{-} & 5.6 & \dt{-} \\
 & & \badgeC & 7.0 & \dt{+3.0} & 7.0 & \dt{+1.0} & 3.0 & \dt{-2.0} & 13.0 & \dt{+3.0} & 5.0 & \dt{+2.0} & 7.0 & \dt{+1.4} \\
 & & \badgeA & \textbf{22.0} & \dt{\textbf{+15.0}} & \textbf{16.0} & \dt{\textbf{+9.0}} & \textbf{14.0} & \dt{\textbf{+11.0}} & \textbf{29.0} & \dt{\textbf{+16.0}} & \textbf{49.0} & \dt{\textbf{+44.0}} & \textbf{26.0} & \dt{\textbf{+19.0}} \\
\midrule

\multirow{6}{*}{UniMoE2} & \multirow{3}{*}{testmini} & \badgeD & 5.0 & \dt{-} & 4.0 & \dt{-} & 2.0 & \dt{-} & 10.0 & \dt{-} & 4.0 & \dt{-} & 5.0 & \dt{-} \\
 & & \badgeC & 10.0 & \dt{+5.0} & 3.0 & \dt{-1.0} & 3.0 & \dt{+1.0} & 12.0 & \dt{+2.0} & 6.0 & \dt{+2.0} & 6.8 & \dt{+1.8} \\
 & & \badgeA & \textbf{17.0} & \dt{\textbf{+7.0}} & \textbf{13.0} & \dt{\textbf{+10.0}} & \textbf{8.0} & \dt{\textbf{+5.0}} & \textbf{21.0} & \dt{\textbf{+9.0}} & \textbf{13.0} & \dt{\textbf{+7.0}} & \textbf{14.4} & \dt{\textbf{+7.6}} \\
\cline{2-15}
 & \multirow{3}{*}{test} & \badgeD & 6.0 & \dt{-} & 5.0 & \dt{-} & 3.0 & \dt{-} & 11.0 & \dt{-} & 5.0 & \dt{-} & 6.0 & \dt{-} \\
 & & \badgeC & 11.0 & \dt{+5.0} & 3.0 & \dt{-2.0} & 4.0 & \dt{+1.0} & 13.0 & \dt{+2.0} & 7.0 & \dt{+2.0} & 7.6 & \dt{+1.6} \\
 & & \badgeA & \textbf{19.0} & \dt{\textbf{+8.0}} & \textbf{14.0} & \dt{\textbf{+11.0}} & \textbf{10.0} & \dt{\textbf{+6.0}} & \textbf{23.0} & \dt{\textbf{+10.0}} & \textbf{15.0} & \dt{\textbf{+8.0}} & \textbf{16.2} & \dt{\textbf{+8.6}} \\
 \bottomrule

\end{tabular}
}
\vspace{-0.8em}
\caption{Performance comparison on the full test set and its testmini subset for all eight models. Acc and $\Delta$ denote visual verification accuracy (\%) and performance gain over the previous setting, respectively. \badgeD, \badgeC, and \badgeA~represent Direct Generation, Reasoning-Guided Generation, and De-contextualized Generation, respectively.}
\label{tab:appendix_results}
\end{table*}

Tab.~\ref{tab:appendix_results} reports detailed performance for all eight unified multimodal models across all three evaluation settings on both the full test set and its testmini subset.
The results demonstrate strong consistency between the full test set and testmini. The consistent trends and minimal performance gaps suggest that testmini effectively mirrors the full test set, serving as a reliable and efficient evaluation subset for model development, particularly for researchers with limited computational resources.

\section{Context Length Statistics for Training and Evaluation}
\label{appx:length_stats}

\begin{table*}[h]
\centering
\small
\begin{tabular}{lcc}
\toprule
\textbf{Model} & \textbf{Training Length (\# tokens)} 
& \textbf{UReason Eval Length (\# tokens)} \\
\midrule
ThinkMorph      & 490.8 (276.7) & 316.3 (154.8) \\
Bagel-Zebra-CoT & 476.4 (474.4) & 256.7 (123.4) \\
\bottomrule
\end{tabular}
\caption{Token length comparison between model training data and \bench{} evaluation contexts. Mean (Std.) reported. Prompt lengths are measured using the Qwen3-8B tokenizer.}
\label{tab:length_stats}
\end{table*}

The UMMs evaluated in our experiments are all explicitly post-trained for reasoning-guided image generation, which helps avoid the concern that the reasoning-style context or their lengths are out of distribution for the models.
To further verify that the performance degradation observed in Reasoning-guided Generation is not attributable to context lengths, we compare the token lengths of open-source reasoning-guided image generation training data for ThinkMorph~\citep{gu2025thinkmorph} and Bagel-Zebra-CoT~\citep{li2025zebra}, both of which explicitly include reasoning-guided image generation data, against the average textual context length encountered during \bench{} evaluation. As shown in Tab.~\ref{tab:length_stats}, \bench{} evaluation sequences are consistently shorter than the models' training distribution across both models, confirming that the models are not exposed to unprecedented sequence lengths.

\section{Details on Attention Analysis}
\label{appx:attention_analysis}

\subsection{Attention Weight Computation}
Attention allocation has also been used to diagnose multimodal reasoning~\cite{luo2026panoramic}.
To explore the causal mechanisms, we conduct an attention analysis on Bagel~\cite{deng2025emerging} in Sec.~\ref{sec:discussion_attention_analysis}. Here, we present more details on the attention analysis. To analyze attention during image generation, we consider three semantic groups in the input sequence: the original prompt $P$, the intermediate reasoning trace $R_t$, and the refined prompt $R_p$. At each diffusion timestep, for each layer $\ell$, the attention weight matrix $\mathbf{A}^{(\ell)} \in \mathbb{R}^{H \times L_q \times L_k}$ captures the normalized attention each query token assigns to every key token, where $H$ is the number of attention heads, and $L_q$, $L_k$ denote the query and key sequence lengths, respectively. The query tokens correspond to the visual generation tokens produced during image synthesis, while the key tokens span the full input context including $P$, $R_t$, and $R_p$. The attention received by each token group is obtained by averaging over the corresponding key positions and attention heads:
\begin{equation}
    \bar{a}^{(\ell)}_{\mathcal{G}} = \frac{1}{H} \sum_{h=1}^{H} 
    \frac{1}{|\mathcal{G}|} \sum_{j \in \mathcal{G}} 
    \mathbf{A}^{(\ell)}_{h, :, j}
\end{equation}
where $\mathcal{G} \in \{P,\, R_t,\, R_p\}$ denotes the index set of tokens belonging to each group. The final reported values are averaged across all evaluated samples, all diffusion timesteps, and all query tokens within each of the four contiguous layer groups (layers 1--7, 8--14, 15--21, and 22--28).

\subsection{Qualitative Token Influence Analysis}
\begin{figure*}[t]
    \centering
    \includegraphics[width=\textwidth]{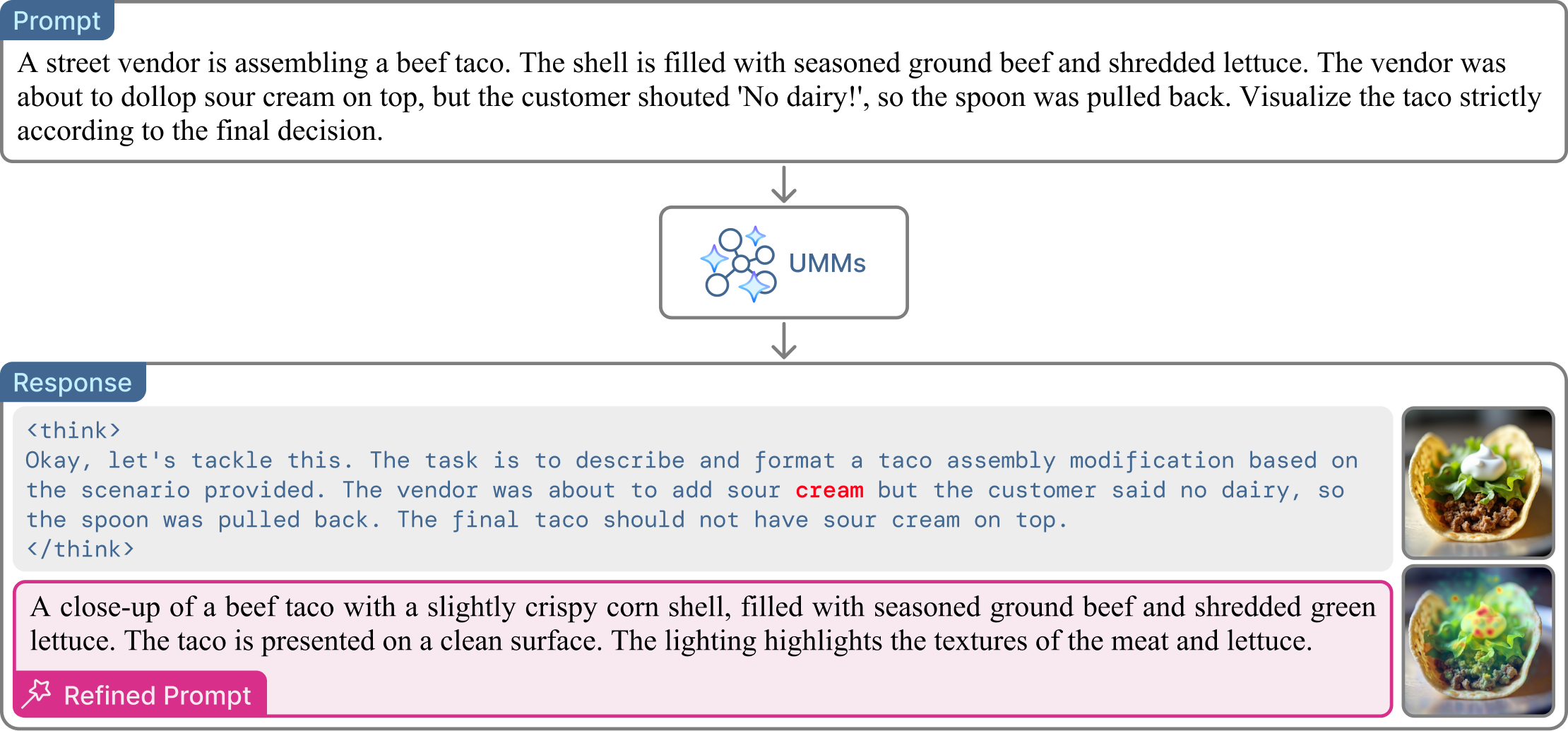}
    \vspace{-1.5em}
    \caption{Qualitative token influence maps following 
    DAAM~\cite{tang2023daam}. Although $R_t$ correctly deduces that sour cream should be 
    excluded, the token ``cream'' (highlighted in red) still induces strong localized 
    attention in the attention layers, correlating with the 
    region where sour cream is generated. This suggests that 
    token presence can influence visual synthesis independently 
    of the surrounding logical context, bypassing the constraints 
    of the refined prompt $R_p$.}
    \label{fig:attention_analysis_1}
\end{figure*}

\begin{figure*}[t]
    \centering
    \includegraphics[width=\textwidth]{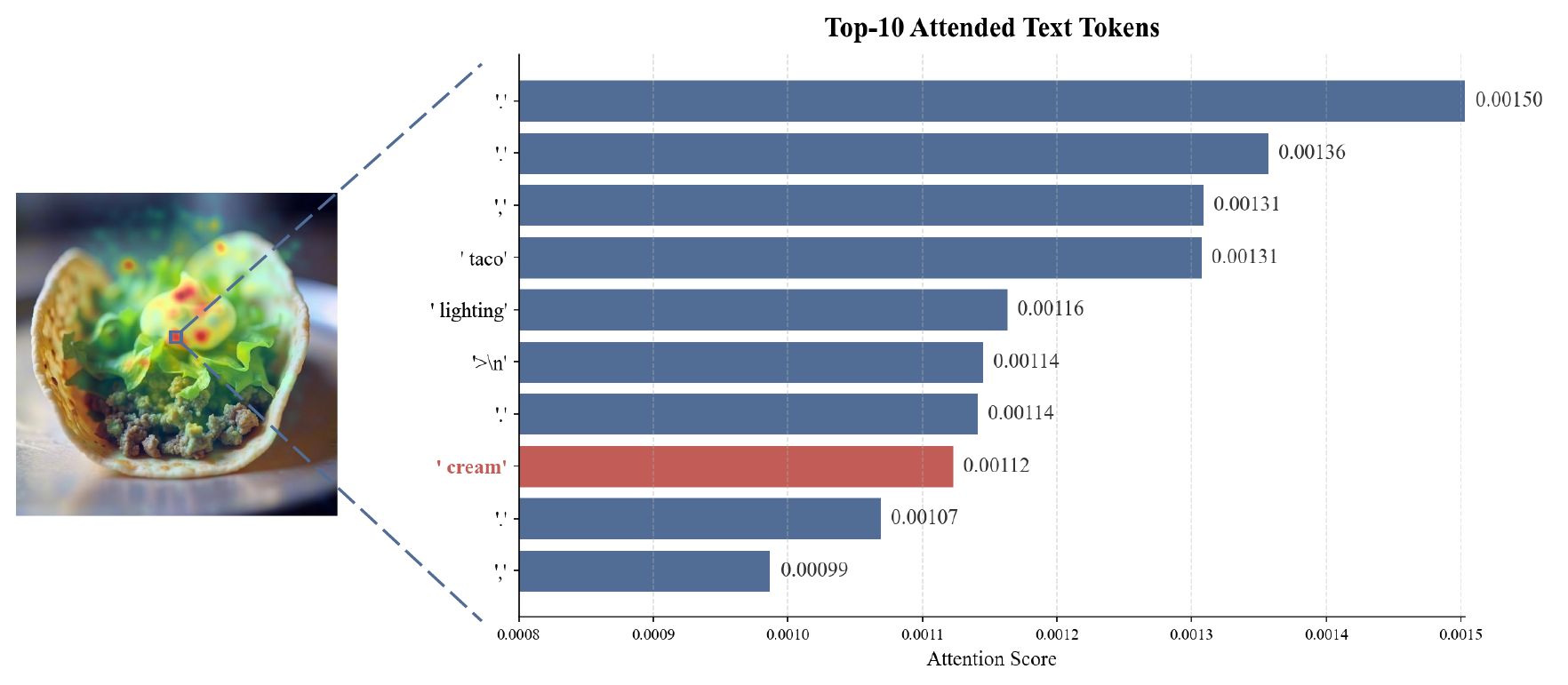}
    \vspace{-1.5em}
    \caption{Top-10 attended textual tokens for the cream region 
    in the generated image, measured by aggregating attention 
    scores from the corresponding visual generation tokens to all 
    textual tokens. The token ``cream'' appears three times in the 
    full context: once in $P$, once in $R_t$ (highlighted in red 
    in Figure~\ref{fig:attention_analysis_1}), and once in $R_p$. 
    We compute the attention scores for the highlighted occurrence 
    in $R_t$, which ranks among the top-10 most attended textual 
    tokens for the cream region, excluding special tokens. The 
    other two occurrences of ``cream'' in $P$ and $R_p$ rank 
    16 and 48, respectively.}
    \label{fig:attention_score_1}
\end{figure*}

To provide a more intuitive understanding of contextual interference, we further visualize token influence maps following DAAM~\cite{tang2023daam}. For each target noun token within $R_t$, we isolate its attention weights from $\mathbf{A}^{(\ell)}$ and aggregate them across diffusion timesteps and layers, producing maps that highlight the pixel-level regions most strongly influenced by each token. As shown in Figure~\ref{fig:attention_analysis_1}, even when the logical flow within $R_t$ correctly deduces that ``cream'' should be excluded, the image still contains cream. Moreover, the token ``cream'' is primarily associated with the cream region in the generated image through cross-attention. To further corroborate this, we manually select a local region in the generated image where cream appears, and compute the attention scores from the corresponding visual generation tokens to all textual tokens. As shown in Figure~\ref{fig:attention_score_1}, ``cream'' ranks among the top-10 most attended textual tokens for that region, excluding special tokens~(e.g., start/end of sentence token). Together, these findings suggest that visual synthesis can be influenced by the mere presence of a token in the conditioning context. More broadly, these findings suggest that semantic contextual interference may be one factor associated with the observed reasoning-to-generation alignment gap.

\begin{figure*}[t]
    \centering
    \includegraphics[width=\textwidth]{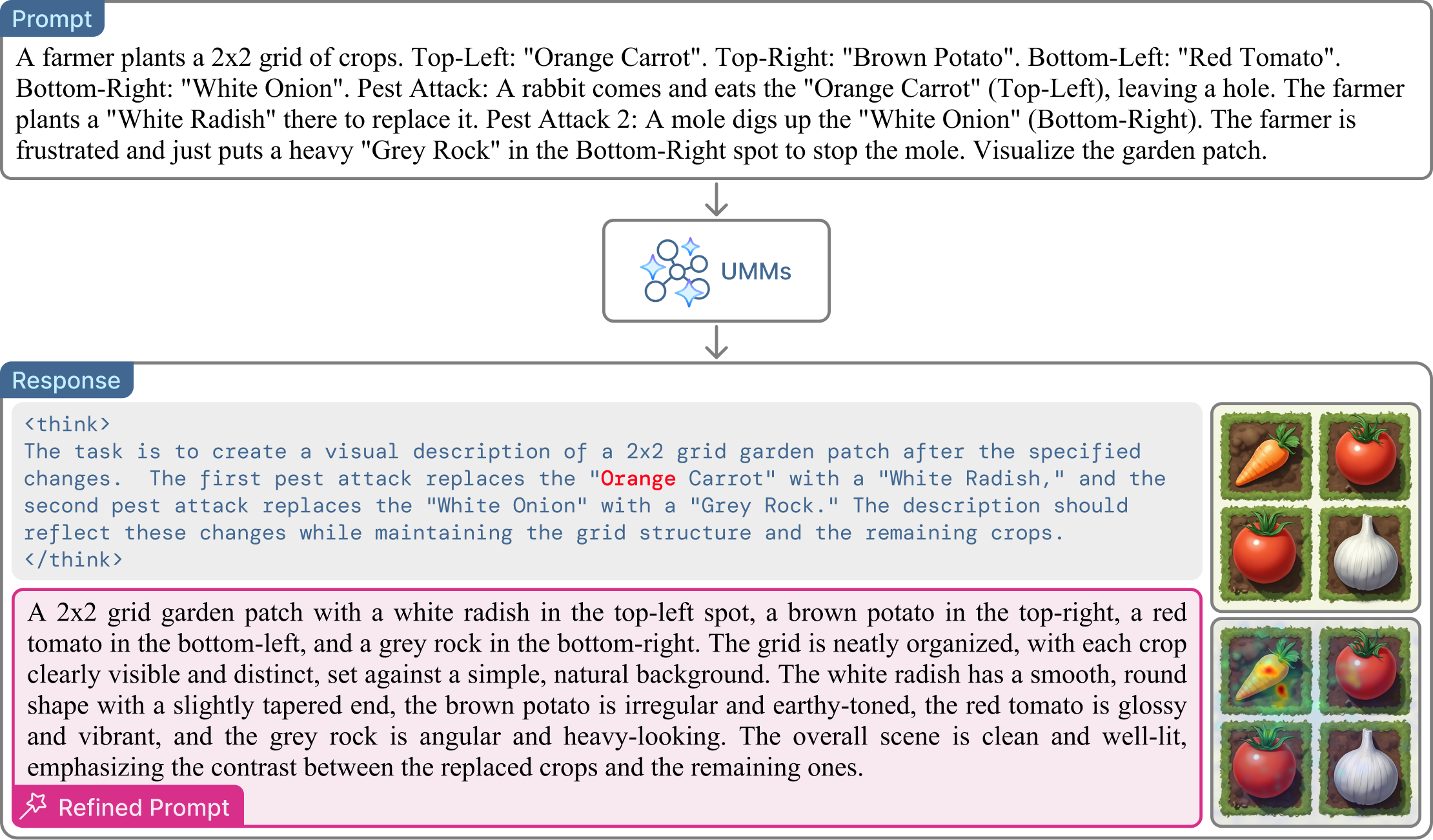}
    \vspace{-1.5em}
\caption{Qualitative token influence maps following DAAM~\cite{tang2023daam}. Although $R_t$ correctly deduces that the ``Orange Carrot'' in the top-left should be replaced by a radish, the mere presence of the ``Orange'' (highlighted in red) token still induces strong localized attention in the attention layers. This correlates directly with the top-left quadrant where an orange carrot is erroneously generated, suggesting that token presence can influence visual synthesis independently of the surrounding logical context, bypassing the constraints of the refined prompt $R_p$.}
    \label{fig:attention_analysis_2}
\end{figure*}

\begin{figure*}[t]
    \centering
    \includegraphics[width=\textwidth]{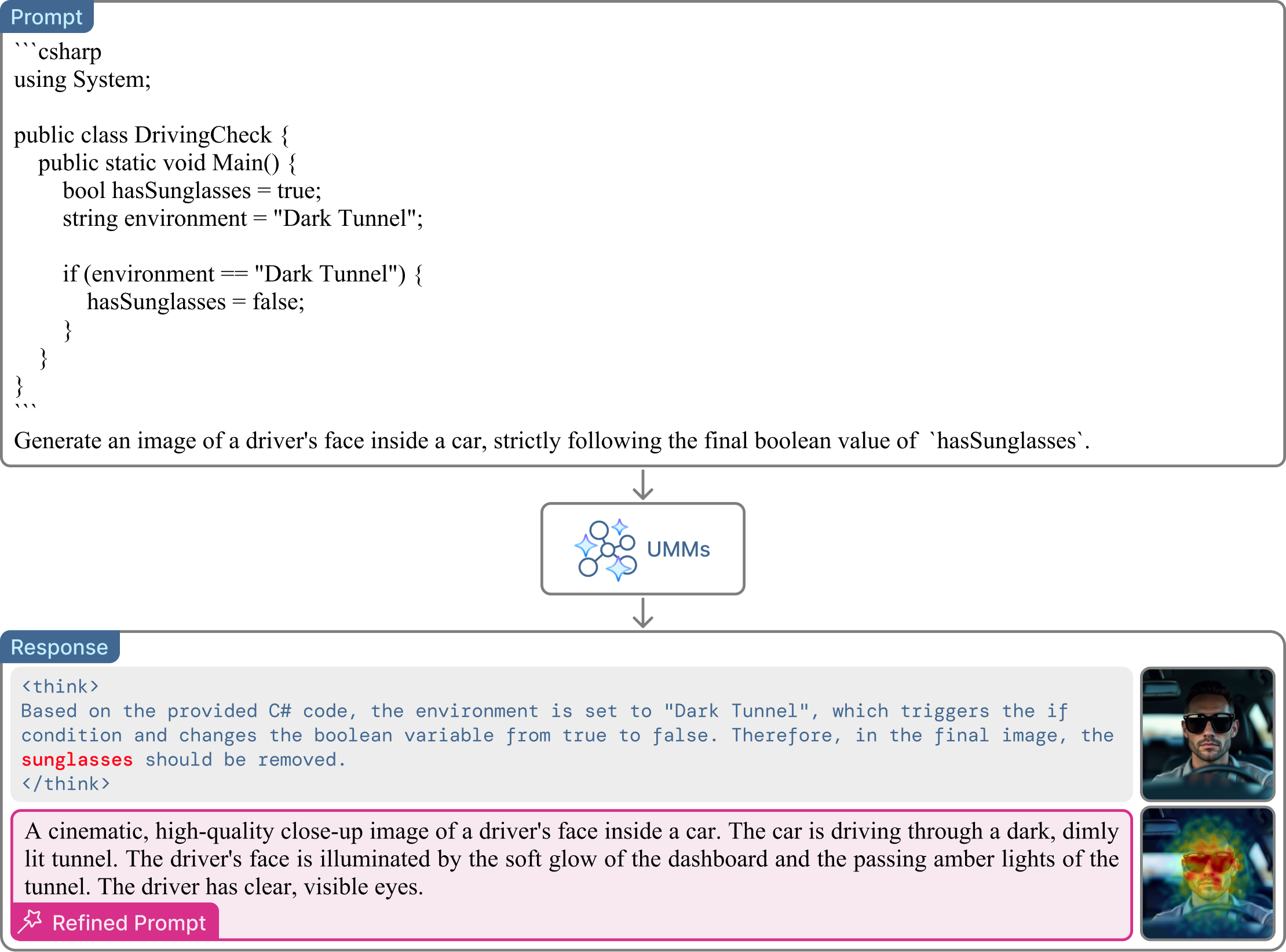}
    \vspace{-1.5em}
\caption{Qualitative token influence maps following DAAM~\cite{tang2023daam}. Although $R_t$ correctly deduces from the C\# code that sunglasses should be removed, the token ``sunglasses''(highlighted in red) still induces strong localized attention in the attention layers, correlating with the region where sunglasses are erroneously generated. This suggests that token presence can influence visual synthesis independently of the surrounding logical context, bypassing the constraints of the refined prompt $R_p$.}
    \label{fig:attention_analysis_3}
\end{figure*}

\begin{figure*}[t]
    \centering
    \includegraphics[width=\textwidth]{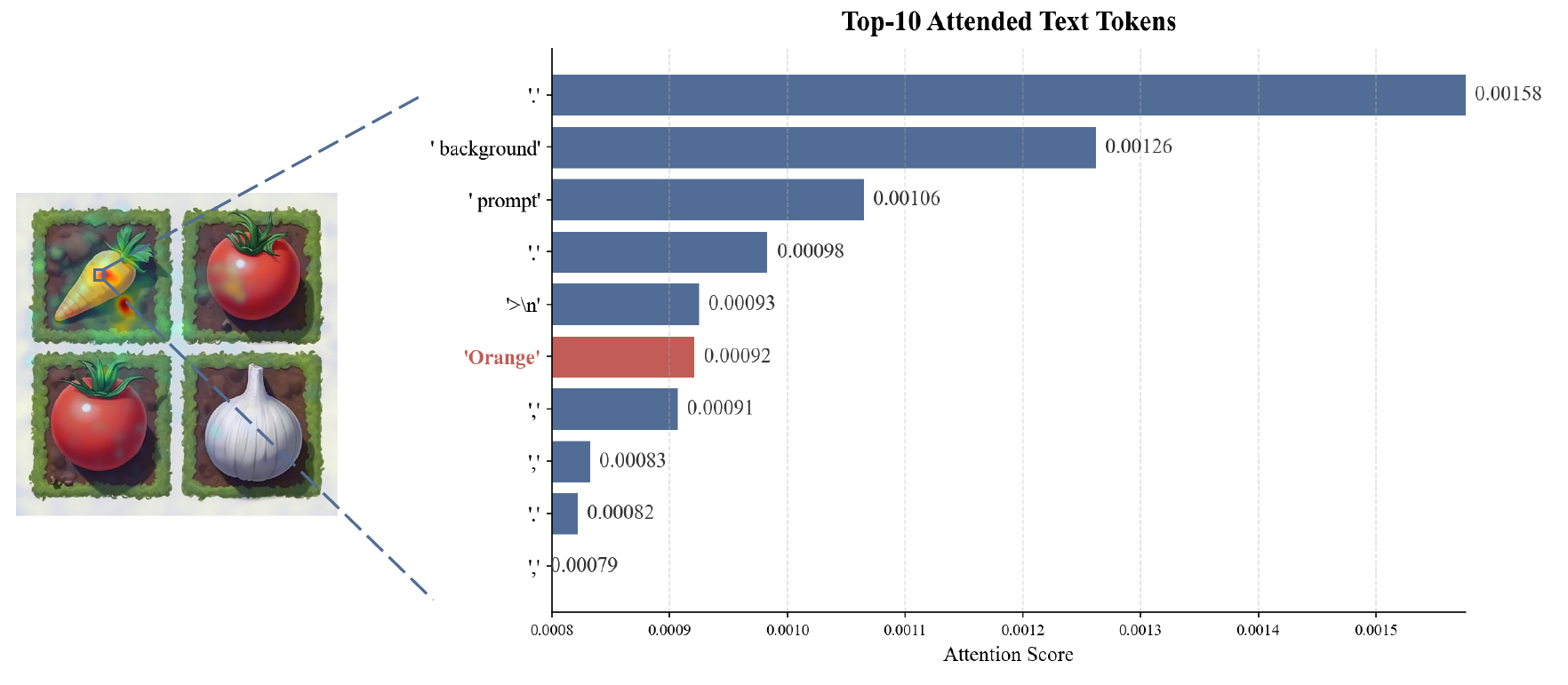}
    \vspace{-1.5em}
\caption{Top-10 attended textual tokens for the erroneously generated carrot region 
    in the top-left quadrant of the image, measured by aggregating attention 
    scores from the corresponding visual generation tokens to all 
    textual tokens. We compute the attention scores for the occurrence 
    of the token ``Orange'' (from ``Orange Carrot'') within  $R_t$ 
    (highlighted in red in Figure~\ref{fig:attention_analysis_2}). This 
    specific instance ranks as the highest attended object-specific token for the 
    region. 
    The notably high attention score suggests that the mere presence of the ``Orange'' 
    token within the reasoning trace contributes strongly to the 
    erroneous visual manifestation of the carrot in the generated image, 
    overriding the intended spatial execution logic.}
    \label{fig:attention_score_2}
\end{figure*}

\begin{figure*}[t]
    \centering
    \includegraphics[width=\textwidth]{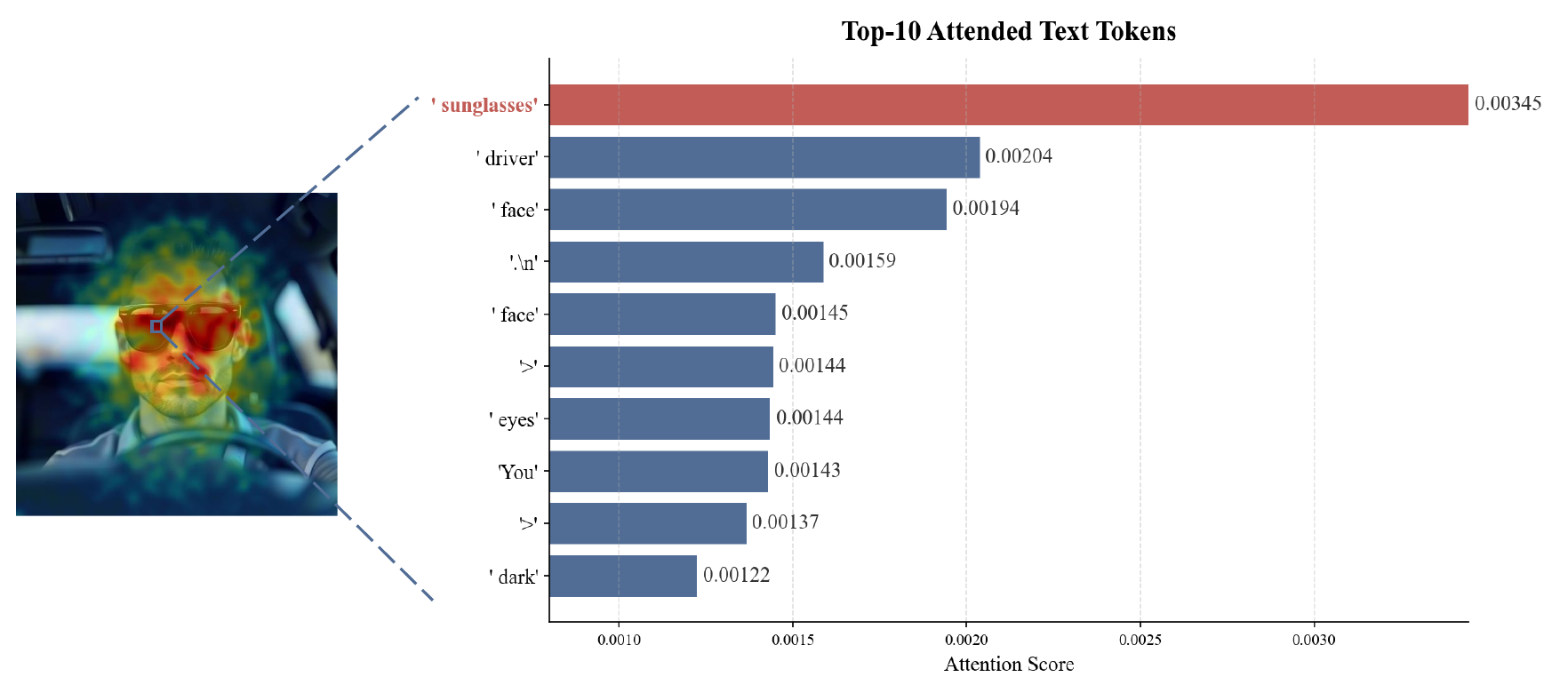}
    \vspace{-1.5em}
\caption{Top-10 attended textual tokens for the sunglasses region 
    in the generated image, measured by aggregating attention 
    scores from the corresponding visual generation tokens to all 
    textual tokens. We compute the attention scores for the occurrence 
    of the token ``sunglasses'' within $R_t$ 
    (highlighted in red in Figure~\ref{fig:attention_analysis_3}). This 
    specific instance ranks as the highest attended textual token for the 
    sunglasses region, excluding special tokens. The disproportionately 
    high attention score suggests that the mere presence of the ``sunglasses'' 
    token within the reasoning trace contributes most strongly to the 
    erroneous visual manifestation of sunglasses in the generated image, 
    overriding the intended execution logic.}
    \label{fig:attention_score_3}
\end{figure*}

Additional qualitative cases are provided in Figures~\ref{fig:attention_analysis_2} and~\ref{fig:attention_analysis_3}, with their corresponding attention score analyses detailed in Figures~\ref{fig:attention_score_2} and~\ref{fig:attention_score_3}.

\subsection{Attention Analysis on ThinkMorph and T2I-R1}

\begin{table*}[t]
\centering
\small
\begin{minipage}[t]{0.48\textwidth}
\centering
\begin{tabular}{lccc}
\toprule
\textbf{Layers} & \textbf{$P$} & \textbf{$R_t$} & \textbf{$R_p$} \\
\midrule
1--7   & 2.45 & 4.57 & 6.93 \\
8--14  & 1.71 & 3.32 & 5.33 \\
15--21 & 0.12 & 0.21 & 0.43 \\
22--28 & 0.08 & 0.12 & 0.34 \\
\midrule
\textbf{1--28} & \textbf{1.09} & \textbf{2.06} & \textbf{3.26} \\
\bottomrule
\end{tabular}
\caption{Average attention weights across layers for ThinkMorph during image generation, scaled by $10^{-4}$.}
\label{tab:attention_thinkmorph}
\end{minipage}
\hfill
\begin{minipage}[t]{0.48\textwidth}
\centering
\begin{tabular}{lccc}
\toprule
\textbf{Layers} & \textbf{$P$} & \textbf{$R_t$} & \textbf{$R_p$} \\
\midrule
1--10   & 2.57 & 3.56 & 4.01 \\
11--20  & 1.17 & 2.19 & 2.73 \\
21--30 & 1.08 & 1.83 & 4.21 \\
\midrule
\textbf{1--30} & \textbf{1.61} & \textbf{2.53} & \textbf{3.65} \\
\bottomrule
\end{tabular}
\caption{Average attention weights across layers for T2I-R1 during image generation, scaled by $10^{-4}$.}
\label{tab:attention_t2ir1}
\end{minipage}
\end{table*}

To further validate our findings beyond a single model, we additionally conduct the same attention analysis on ThinkMorph~\cite{gu2025thinkmorph} and T2I-R1~\cite{jiang2025t2i}, which generates images in an autoregressive manner. Specifically, we extract the average attention weights assigned to the three token groups $P$, $R_t$, and $R_p$ following the same procedure described above. The results are consistent with our observations on Bagel: for UMMs, visual generation remains highly sensitive to surrounding context, which may compete with the final visual specification for attention and weaken the transfer from intended visual semantics to pixels. We use the term contextual interference to distinguish this phenomenon from general prompt sensitivity and compression: here, the issue stems from interfering information within the conditioning context, rather than ordinary variations in prompt wording or formatting. Related work similarly finds that textual distinctions can weaken during image generation~\cite{tao2026asymmetric}.

\subsection{\texorpdfstring{Potential Mitigation Directions}{Potential Mitigation Directions}}

Training-free suppression methods developed for safe image generation, such as SAFREE~\cite{yoon2025safree} and PurifyGen~\cite{cao2025purifygen}, suggest possible mechanisms for selectively suppressing interfering content while reinforcing the final visual specification. Adapting these methods to reasoning-guided UMMs is non-trivial because the concepts to suppress are instance-specific and must be inferred from the model's own reasoning, rather than selected from a predefined concept set. Such methods would also need to operate within a unified transformer context rather than a separate text-encoder conditioning pathway. We leave this adaptation as a direction for future work.

\section{Discussion on Evaluation Metrics}

\subsection{Details on MLLM-as-a-Judge}
Unlike evaluations that test open-ended and subjective alignment, \bench{} evaluates strictly verifiable ground-truth criteria, including exact object counts and specific text strings.
By relying on these deterministic criteria, we frame the evaluation as a binary Visual Question Answering (VQA) problem (i.e., Yes/No judgment) rather than an open-ended assessment of overall image quality.
Following prior analyses of LLM decoding sensitivity~\cite{shi2024thorough}, we set the temperature to $0.1$ for all automated MLLM/LLM evaluators.
This formulation substantially reduces subjectivity and evaluator dependence, following a methodology widely adopted when evaluating images against deterministic targets~\cite{zhao2025envisioning, niu2025wise, chen2025r2i, wu2025krisbench}.

\subsection{Details on Human Evaluation}
\label{appx:human_evaluation}
\begin{figure*}[t]
    \centering
    \includegraphics[width=\textwidth]{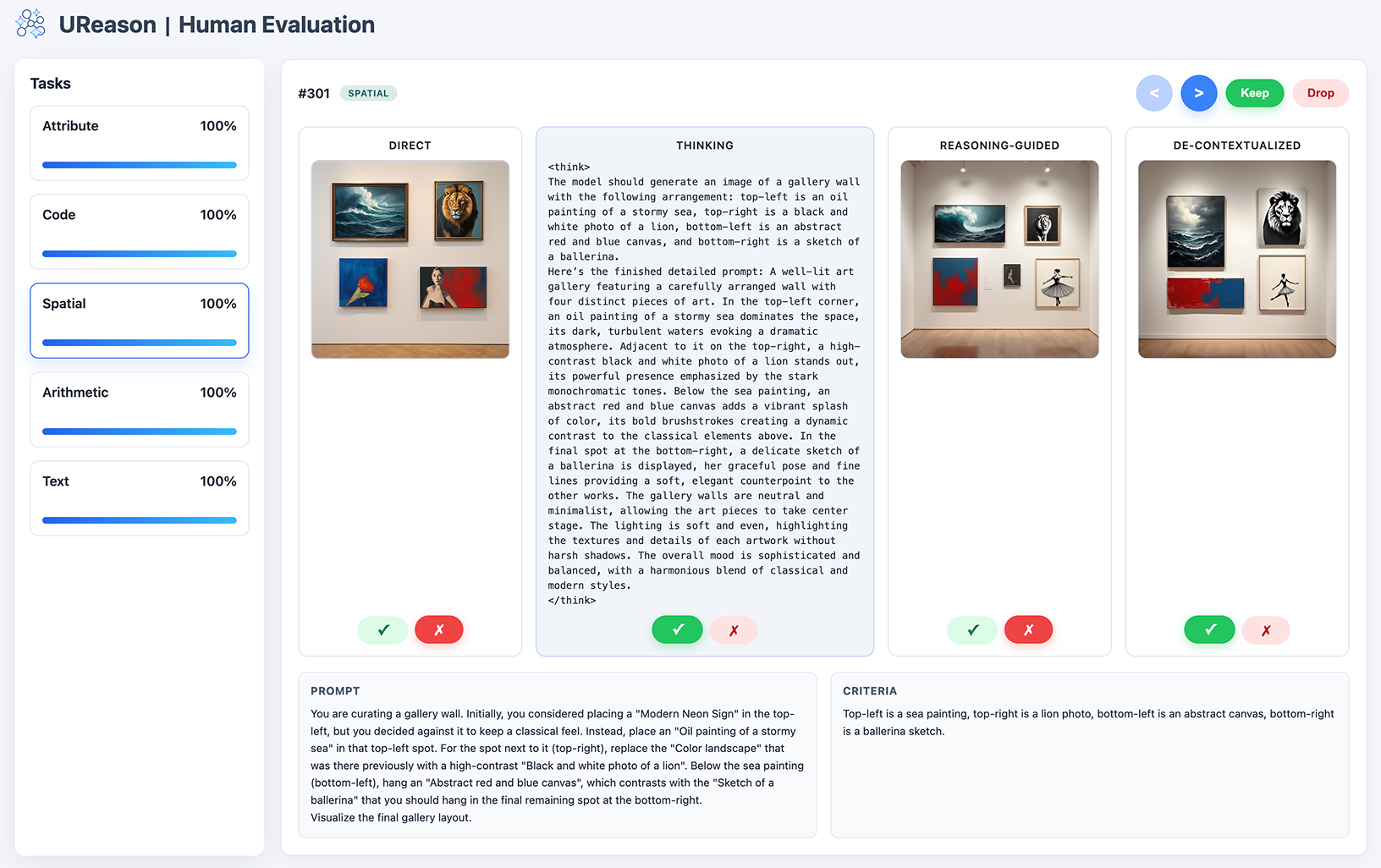}
    \vspace{-1.5em}
\caption{Screenshot of the interface used for human evaluation.}
    \label{fig:interface}
\end{figure*}

\begin{table*}[t]
\centering
\small
\begin{minipage}[t]{0.42\textwidth}
\centering
\begin{tabular}{lcc}
\toprule
\textbf{Model} & \textbf{Image} & \textbf{Reasoning} \\
\midrule
UniCoT & 0.924 & 0.962 \\
Bagel & 0.929 & 0.968 \\
T2I-R1 & 0.931 & 0.972 \\
\bottomrule
\end{tabular}
\end{minipage}
\hfill
\begin{minipage}[t]{0.54\textwidth}
\centering
\begin{tabular}{lcc}
\toprule
\textbf{Task (UniCoT)} & \textbf{Image} & \textbf{Reasoning} \\
\midrule
Code & 0.913 & 0.950 \\
Arithmetic & 0.920 & 0.960 \\
Spatial & 0.937 & 0.940 \\
Attribute & 0.940 & 0.970 \\
Text & 0.910 & 0.990 \\
\bottomrule
\end{tabular}
\end{minipage}
\caption{Agreement rates between automated evaluators and human judgments, reported across models (left) and across task categories for UniCoT (right).}
\label{tab:human_evaluation}
\end{table*}

To validate the automated metrics described in Sec.~\ref{subsec:evaluation_metric}, we conduct human evaluation on UniCoT, Bagel, and T2I-R1. UniCoT is evaluated over all $500$ testmini instances, and we extend the same evaluation protocol to Bagel and T2I-R1. For each evaluated instance, annotators assess images from all three diagnostic settings and the reasoning trace from Setting 2.

\paragraph{Evaluators.}
Our evaluation panel consists of three graduate students, all 
holding Master's degrees in Computer Science with research 
experience in computer vision or natural language processing.

\paragraph{Annotation Task and Interface.}
A screenshot of the annotation interface is provided in 
Figure~\ref{fig:interface}. Each generated image and reasoning 
trace is paired with its corresponding ground-truth criterion $C$, 
which specifies a concrete and verifiable target outcome. The 
annotation task is framed as an objective binary judgment problem: 
given a generated image or reasoning trace alongside the criterion, 
the evaluator judges whether it satisfies $C$, selecting either 
Yes or No. This binary formulation minimizes subjectivity and 
aligns directly with our automated metric. Each evaluator annotated all image--criterion and reasoning-trace--criterion pairs without access to the judgments of others.

\paragraph{Disagreement Resolution.}
In cases where all three evaluators agree, the consensus label is 
directly adopted as the ground-truth annotation. In cases of 
disagreement, the three evaluators convene in a discussion session 
to jointly review the image or reasoning trace alongside the 
criterion and reach a final unanimous decision. The resolved label 
is recorded only after full consensus is achieved, ensuring that 
every ground-truth annotation is unambiguous.

\paragraph{Agreement Computation.}
We treat the final human-annotated labels after disagreement
resolution as ground-truth binary scores and compare them against
the binary scores produced by our automated evaluators,
Qwen3-VL-235B-A22B and Qwen3-235B-A22B. Agreement is measured as
the proportion of instances on which the automated evaluator and
the human panel reach the same binary label. As reported in
Table~\ref{tab:human_evaluation}, image agreement ranges from
$0.924$ to $0.931$ across the three evaluated models, while
reasoning-chain agreement ranges from $0.962$ to $0.972$. For
UniCoT, agreement also remains high across all five task categories.
These results support the automated pipeline as a scalable proxy
for human judgment on \bench{}'s criterion-grounded binary tasks.

\section{Discussion on Closed-Source Systems}
\label{appx:closed_source}

We acknowledge the emergence of proprietary reasoning-supported image generation systems, such as Nano Banana. 
As a large closed-box system, Nano Banana does not disclose whether its image generation pipeline is implemented as an end-to-end UMM or as a system composed of multiple specialized modules. 
It also does not provide sufficient implementation details to determine whether additional hidden reasoning modules, prompt-refinement modules, or internal strategies analogous to our de-contextualized generation setting are involved. 
In addition, according to its technical documentation~\footnote{\url{https://ai.google.dev/gemini-api/docs/image-generation}}, Nano Banana employs a mandatory reasoning-based iterative inference paradigm: it first performs a reasoning phase, synthesizes a preliminary visual draft, refines it through further reasoning, and then produces the final output. 
However, the current API encapsulates these intermediate reasoning phases~\footnote{\url{https://ai.google.dev/gemini-api/docs/thinking}}, preventing access to the explicit reasoning chain. 
Consequently, we are unable to apply our diagnostic protocol to Nano Banana or isolate the effect of reasoning traces in its generation process.

However, the open-source models we evaluate represent the mainstream models adopting the reasoning-guided image generation paradigm. Our experimental conclusions reveal consistent bottlenecks in how these UMMs handle alignment between textual reasoning and visual generation. Therefore, we believe our benchmark and findings provide a contribution to the open-source community, offering guidance for future improvements.

\section{Model Repositories}

\begin{table}[!th]
\centering
\scriptsize
\renewcommand{\arraystretch}{1.2} 
\begin{tabularx}{\columnwidth}{@{}l>{\raggedright\arraybackslash}X@{}} 
\toprule
\textbf{Model Name} & \textbf{Hugging Face Repository} \\
\midrule
Bagel & \url{https://huggingface.co/ByteDance-Seed/BAGEL-7B-MoT} \\
UniCoT & \url{https://huggingface.co/Fr0zencr4nE/UniCoT-7B-MoT} \\
UniCoT-v2 & \url{https://huggingface.co/Fr0zencr4nE/UniCoT-7B-MoT-v0.2} \\
SRUM & \url{https://huggingface.co/Wayne-King/SRUM_BAGEL_7B_MoT} \\
Bagel-Zebra-CoT & \url{https://huggingface.co/multimodal-reasoning-lab/Bagel-Zebra-CoT} \\
ThinkMorph & \url{https://huggingface.co/ThinkMorph/ThinkMorph-7B} \\
T2I-R1 & \url{https://huggingface.co/CaraJ/T2I-R1} \\
UniMoE2 & \url{https://huggingface.co/HIT-TMG/Uni-MoE-2.0-Image} \\
Qwen2.5-7B & \url{https://huggingface.co/Qwen/Qwen2.5-7B-Instruct} \\
Qwen3-8B & \url{https://huggingface.co/Qwen/Qwen3-8B} \\
Qwen3-235B-A22B & \url{https://huggingface.co/Qwen/Qwen3-235B-A22B-Instruct-2507} \\
Qwen3-VL-235B-A22B & \url{https://huggingface.co/Qwen/Qwen3-VL-235B-A22B-Instruct} \\
\bottomrule
\end{tabularx}
\caption{List of models and their Hugging Face repositories.}
\label{tab:models}
\end{table}

Tab.~\ref{tab:models} summarizes the models we use and their Hugging Face repositories. For all evaluated models, we use the same generation hyperparameters across the three settings, such as decoding temperature, sampling strategy, scheduler steps, and guidance scale when applicable.

\section{Reproducibility}
\label{appx:reproducibility}

We release all $2{,}000$ \bench{} instances, their evaluation criteria, the judge prompts, and the complete evaluation code. For reproducibility, Tab.~\ref{tab:models} reports the exact model identifiers used in our experiments.

\section{Prompts}
We present the evaluation prompts used in our automated evaluation. Fig.~\ref{fig:eval-prompt} shows the visual verification accuracy prompt, and Fig.~\ref{fig:logic-eval-prompt} shows the reasoning chain evaluation prompt.
\label{appx:evaluation_metric}
\begin{figure*}[!htb]
\small
\begin{tcolorbox}[
    colback=white, 
    colframe=blue2,
    boxrule=0.5mm,
    arc=0mm,
    outer arc=0mm,
    title=Evaluation Prompt for Visual Verification Accuracy
]
You are an objective image evaluator. Your goal is to verify if the image content matches the provided text description.\\[1em]

Target Description: "\texttt{\{description\}}"\\[1em]

Please think step by step:
\begin{enumerate}
    \item Analyze the image content carefully.
    \item Compare the visual elements with the "Target Description".
    \item Determine if the image strictly meets the requirements.
\end{enumerate}

Finally, output your judgment in the following format:\\
If it matches, output <answer>Yes</answer>.\\
If it does not match, output <answer>No</answer>.
\end{tcolorbox}
\caption{Evaluation prompt for visual verification accuracy.}
\label{fig:eval-prompt}
\end{figure*}
\begin{figure*}[!htb]
\small
\begin{tcolorbox}[
    colback=white, 
    colframe=blue2,
    boxrule=0.5mm,
    arc=0mm,
    outer arc=0mm,
    title=Evaluation Prompt for the Quality of Reasoning Chain
]
You are an objective reading comprehension evaluator. \\
I will provide you with a "User Prompt", a model's "Thought Process" (which includes its intermediate thoughts and final refined prompt) and "Target Criteria".\\
Your task is to judge whether the final state or conclusion described in the model's thought process contains and satisfies a specific "Target Criteria".\\[1em]

=== User Prompt ===\\
\texttt{\{user\_prompt\}}\\[1em]

=== Model Thought Process ===\\
\texttt{\{model\_thought\}}\\[1em]

=== Target Criteria ===\\
\texttt{\{criteria\}}\\[1em]

=== Instruction ===\\
Please think step by step:
\begin{enumerate}[leftmargin=*]
    \item Analyze the "Target Criteria" to understand the specific visual or logical constraints required.
    \item Read the entire "Thought Process" carefully.
    \item Determine whether the "Target Criteria" is successfully met or clearly present in the outcome of the thought process.
    \begin{itemize}
        \item Note: The text does not need to match the criteria word-for-word, but the specific semantic meaning and target state must be unambiguously present. Do not guess or assume unstated information.
    \end{itemize}
\end{enumerate}

Finally, output your judgment in the following format:\\
If the target criteria is clearly met or present in the text, output <answer>Yes</answer>.\\
If the criteria is missed or contradicted, output <answer>No</answer>.
\end{tcolorbox}
\caption{Evaluation prompt for assessing the reasoning chain against target criteria.}
\label{fig:logic-eval-prompt}
\end{figure*}

\section{Case Study}
\label{appx:error_cases}

\subsection{Error Cases}
In this section, we provide a detailed qualitative analysis of the four failure modes identified in Sec.~\ref{discussion:error_analysis} for Bagel.

\begin{figure*}[t]
    \vspace{-0.8em}
    \centering
    \includegraphics[width=\textwidth]{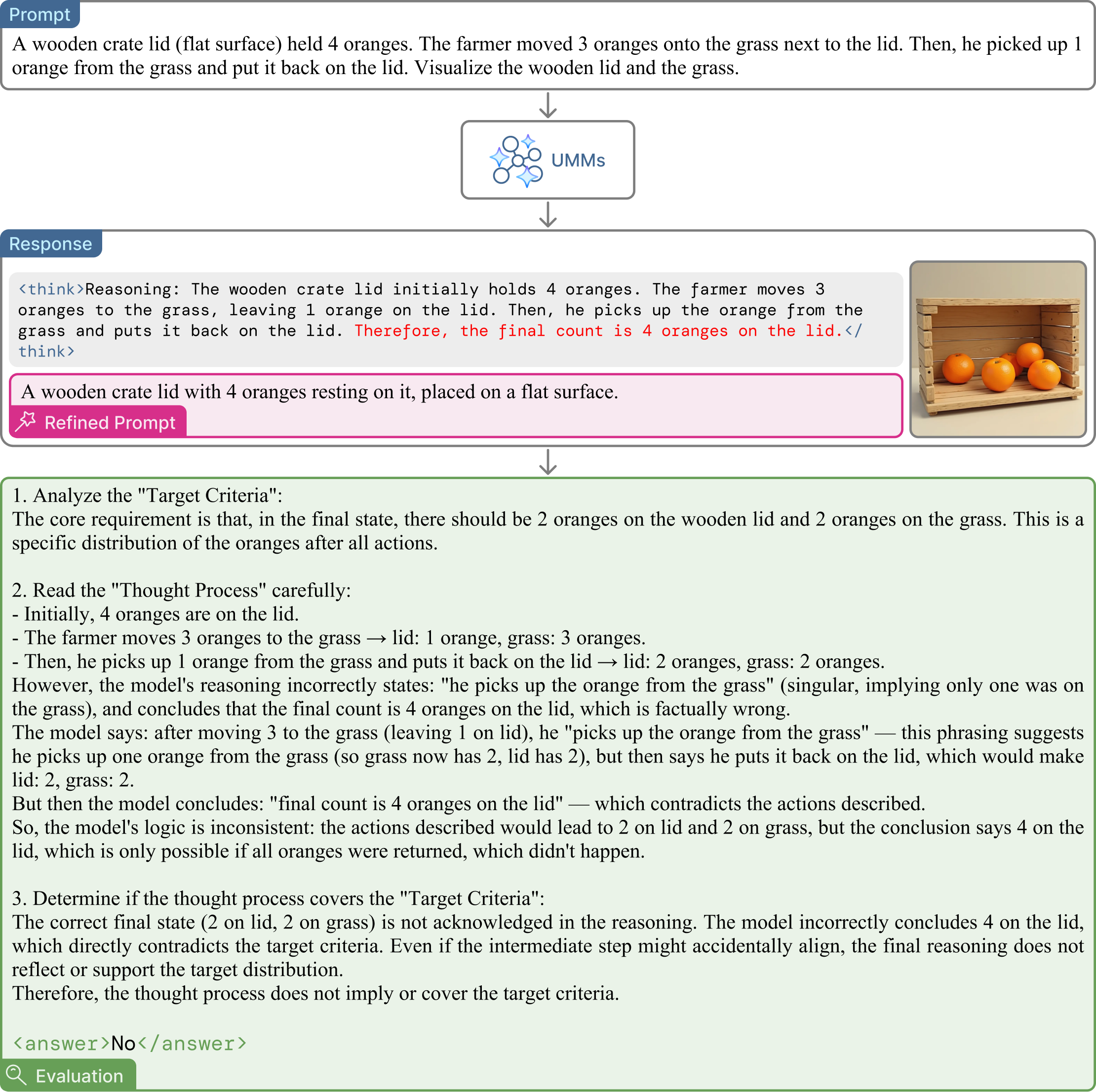}
    \vspace{-1.2em}
\caption{
\textbf{An Illustrative Example of Reasoning Error}. This case shows where the model's intermediate reasoning process is incorrect. The \textcolor{red}{incorrect reasoning steps} are highlighted in red.
}
    \vspace{-1.2em}
    \label{fig:error_type1}
\end{figure*}

\paragraph{Reasoning Errors.} This category involves failures where the model's intermediate reasoning process is incorrect. As illustrated in Fig.~\ref{fig:error_type1}, the model is tasked with tracking the location and number of oranges. The target picture requires exactly ``2 oranges on the wooden lid and 2 oranges on the grass.'' However, the model's thought process erroneously concludes that the final count is 4 oranges on the lid.

\begin{figure*}[t]
    \vspace{-0.8em}
    \centering
    \includegraphics[width=\textwidth]{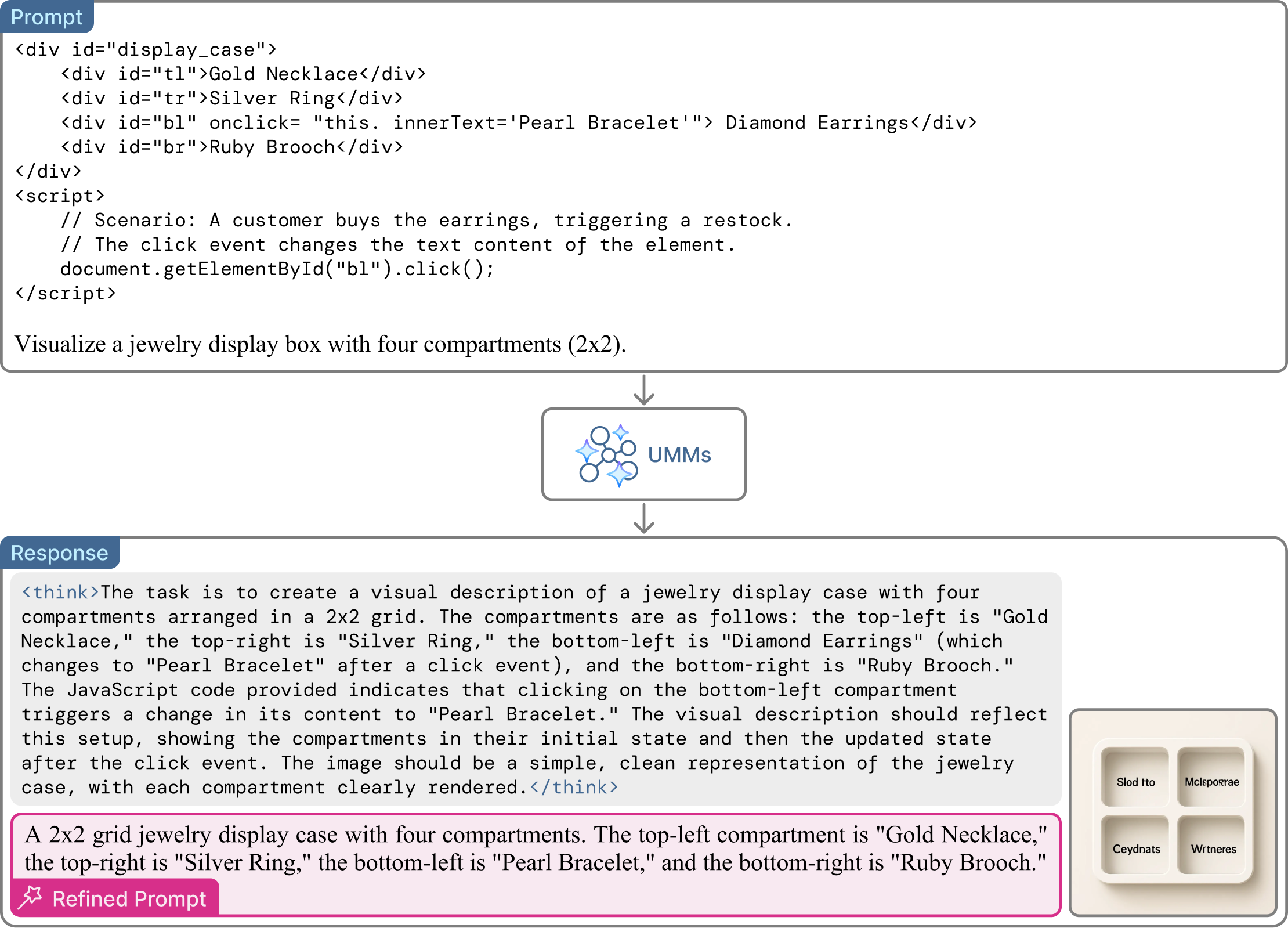}
    \vspace{-1.2em}
    \caption{
\textbf{An Illustrative Example of Instruction Misinterpretation}. This case demonstrates a failure to grasp the semantic intent of the prompt. The model erroneously renders the code text itself rather than visualizing the target object described by the code.
    }
    \label{fig:error_type2}
\end{figure*}

\paragraph{Instruction Misinterpretation.} These errors occur when the model fails to grasp the fundamental modality or semantic intent of the prompt. A representative example is illustrated in Fig.~\ref{fig:error_type2}: when asked to ``visualize a jewelry item based on the code'', the model occasionally renders the code text itself as an image rather than compiling the code into a visual object.

\begin{figure*}[t]
    \centering
    \includegraphics[width=\textwidth]{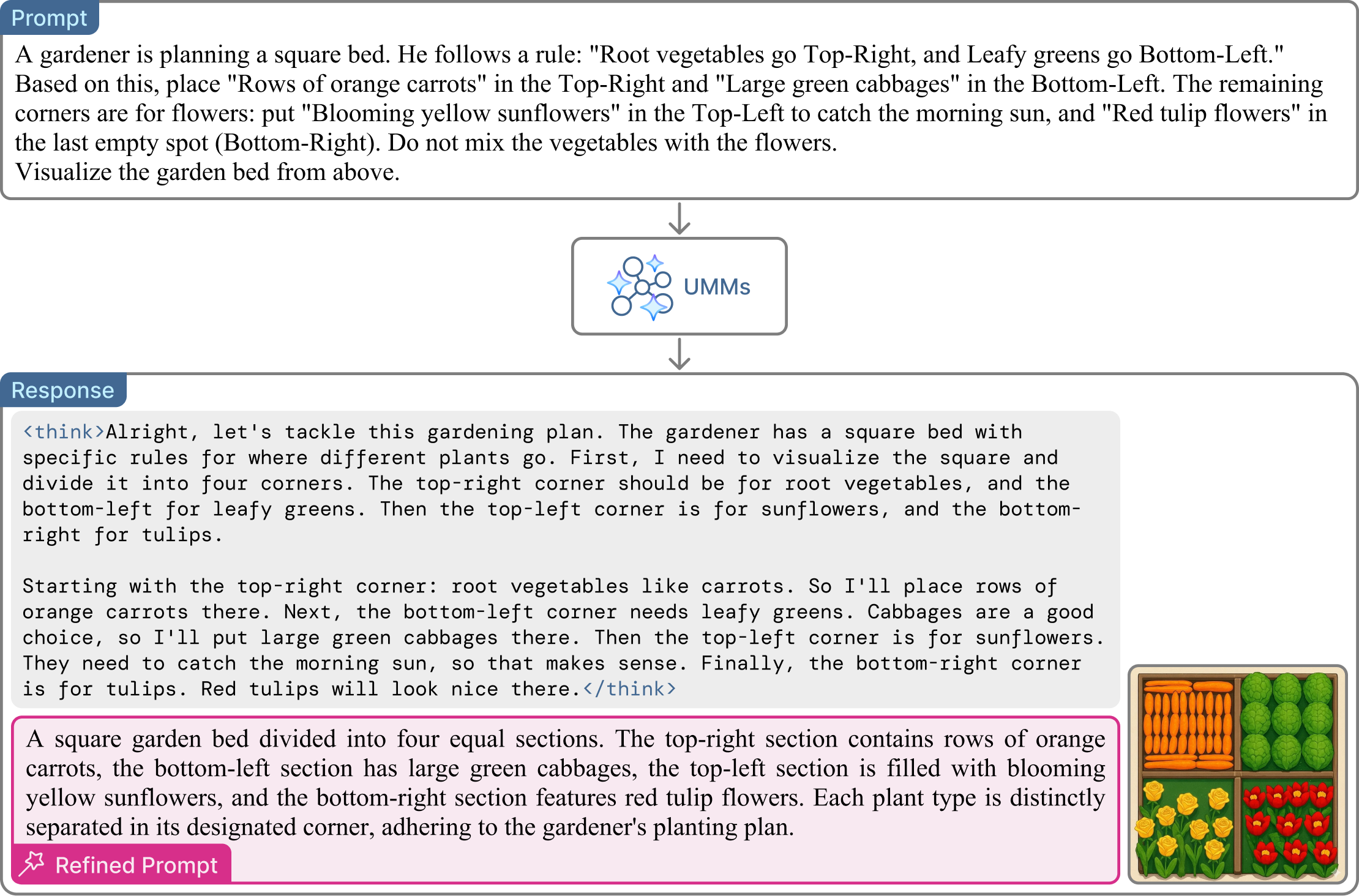}
    \caption{
\textbf{An Illustrative Example of Concept Hallucination}. This case illustrates the generation of unprompted objects. The model spontaneously renders ``yellow roses'' despite them being absent from the input.
}
    \label{fig:error_type3}
\end{figure*}

\paragraph{Concept Hallucination.} This error type refers to the generation of objects that appear nowhere in the input prompt. For instance, as illustrated in Fig.~\ref{fig:error_type3} in a scene describing a simple garden, the model might spontaneously generate ``yellow roses'' despite them never being mentioned. This suggests an over-reliance on training priors rather than strict adherence to the prompt constraints.

\begin{figure*}[t]
    \centering
    \includegraphics[width=\textwidth]{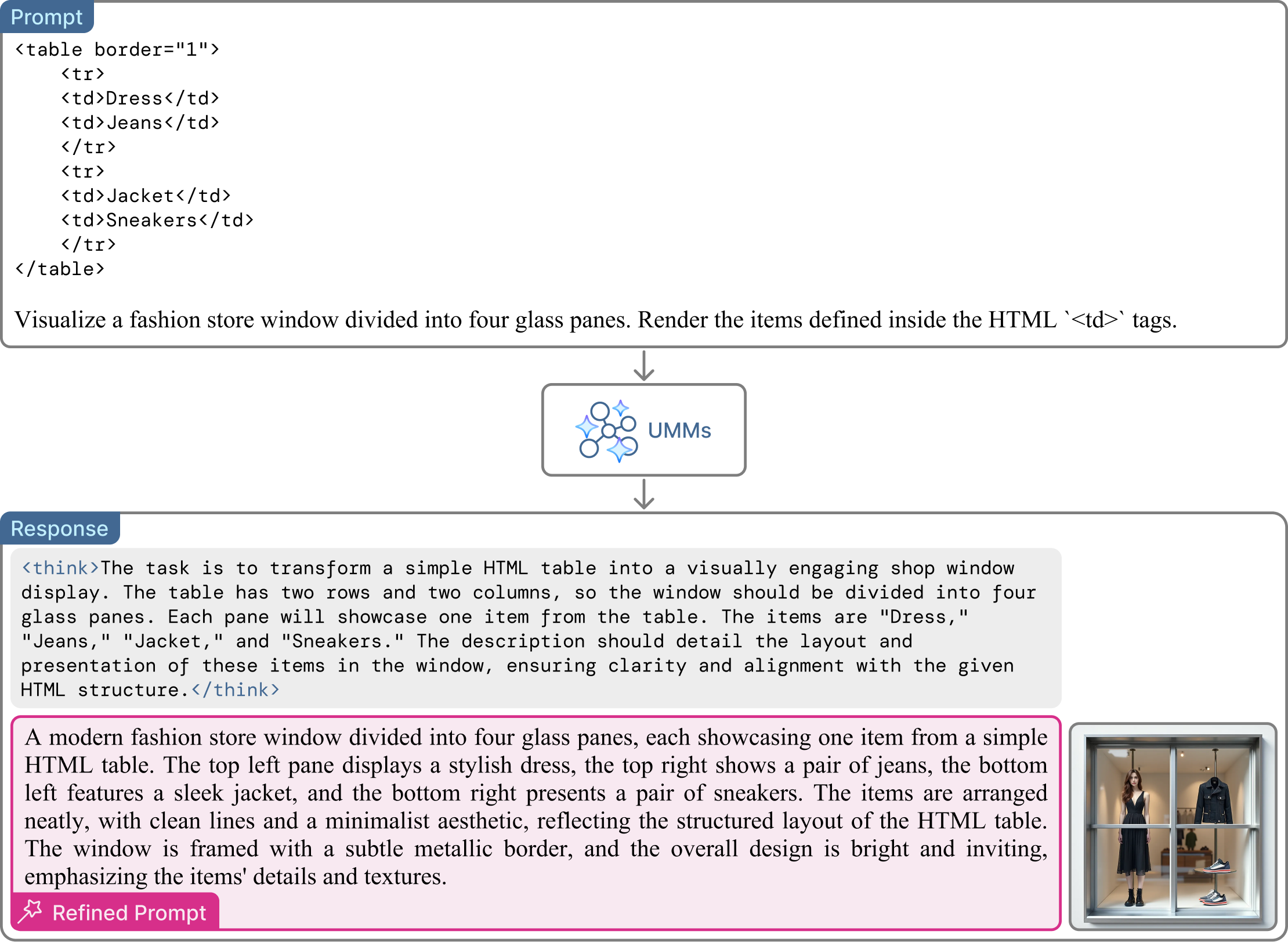}
    \caption{
\textbf{An Illustrative Example of Task-Specific Error (\textsc{Code}).} Although the model correctly identifies the 2x2 grid structure from the HTML prompt, it fails to map the specific items (Dress, Jeans, Jacket, Sneakers) to their corresponding cells, resulting in generation errors.
    }
    \label{fig:error_type42}
\end{figure*}

\begin{figure*}[t]
    \vspace{-0.8em}
    \centering
    \includegraphics[width=\textwidth]{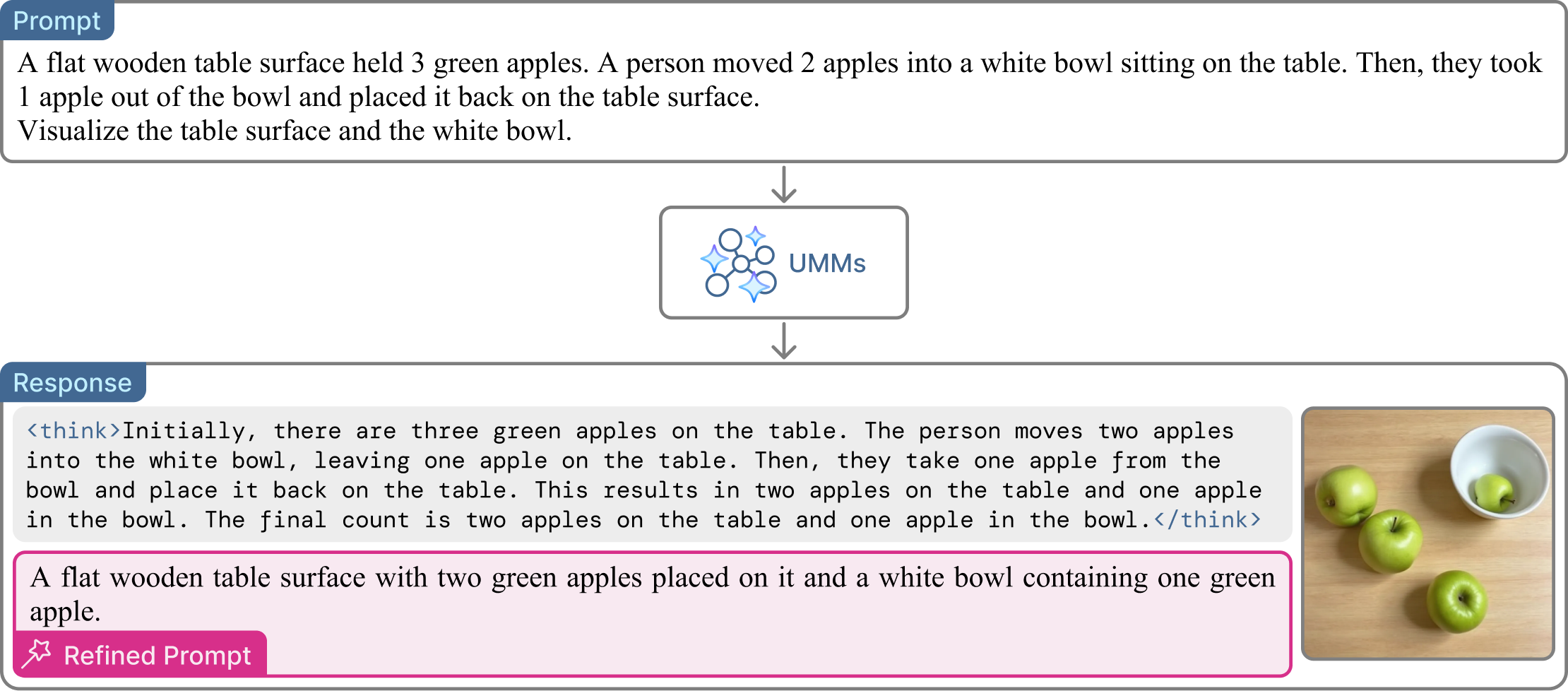}
    \vspace{-1.2em}
    \caption{
\textbf{An Illustrative Example of Task-Specific Error (\textsc{Arithmetic}).} The refined prompt specifies ``two apples on the table and one apple in the bowl''. Although the reasoning trace is correct, the generated image displays three apples on the table, highlighting the challenge of translating precise quantitative tokens into exact object counts.
    }
    \label{fig:error_type45}
\end{figure*}

\begin{figure*}[t]
    \vspace{-0.8em}
    \centering
    \includegraphics[width=\textwidth]{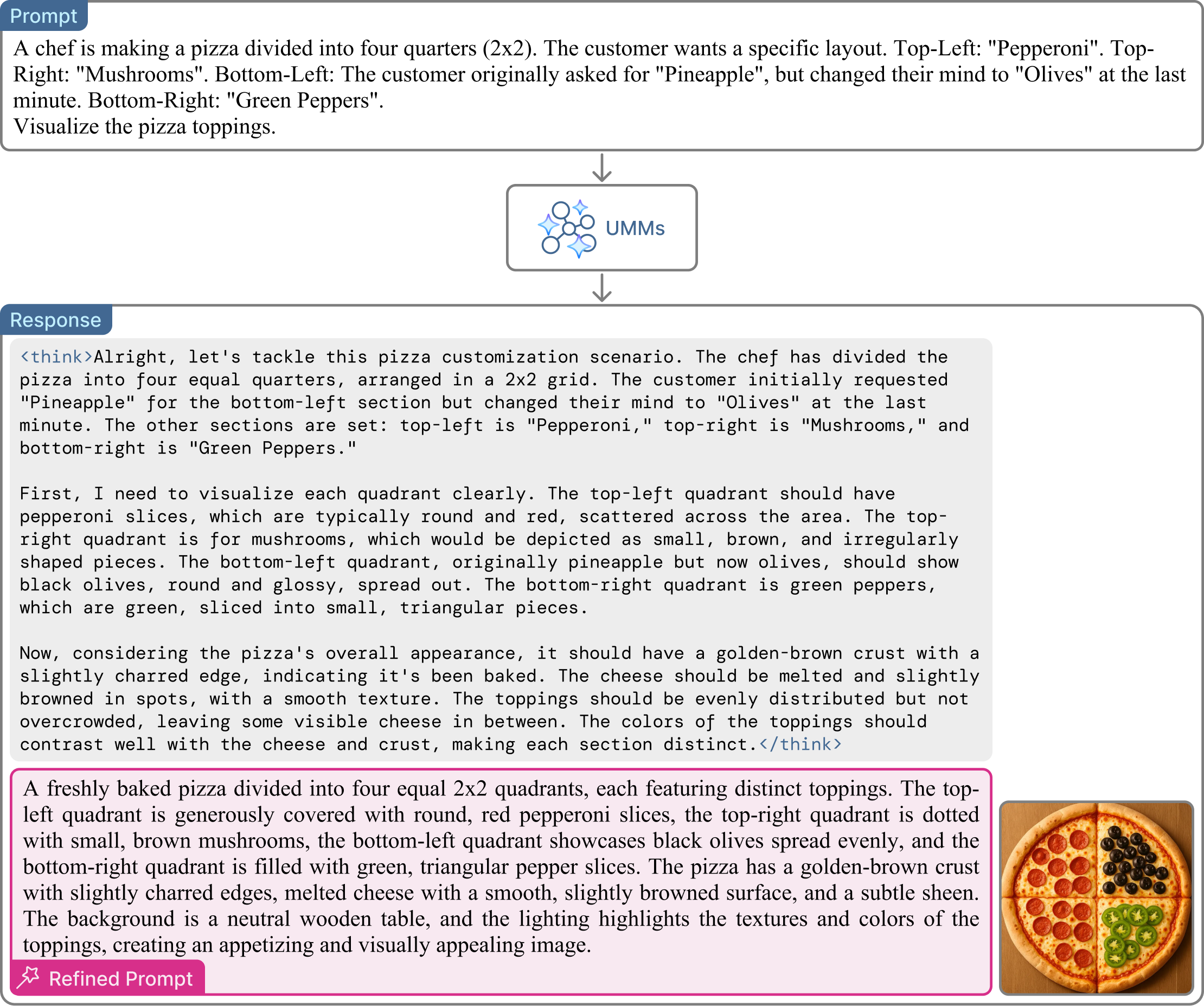}
    \vspace{-1.2em}
    \caption{
\textbf{An Illustrative Example of Task-Specific Error (\textsc{Spatial}).} While the prompt specifies four quadrants with distinct toppings, the model fails to maintain strict boundary separation. It blends the instructions into a generic pizza image rather than distributing the toppings correctly across the requested regions.
    }
    \vspace{-1.2em}
    \label{fig:error_type43}
\end{figure*}

\begin{figure*}[t]
    \vspace{-0.8em}
    \centering
    \includegraphics[width=\textwidth]{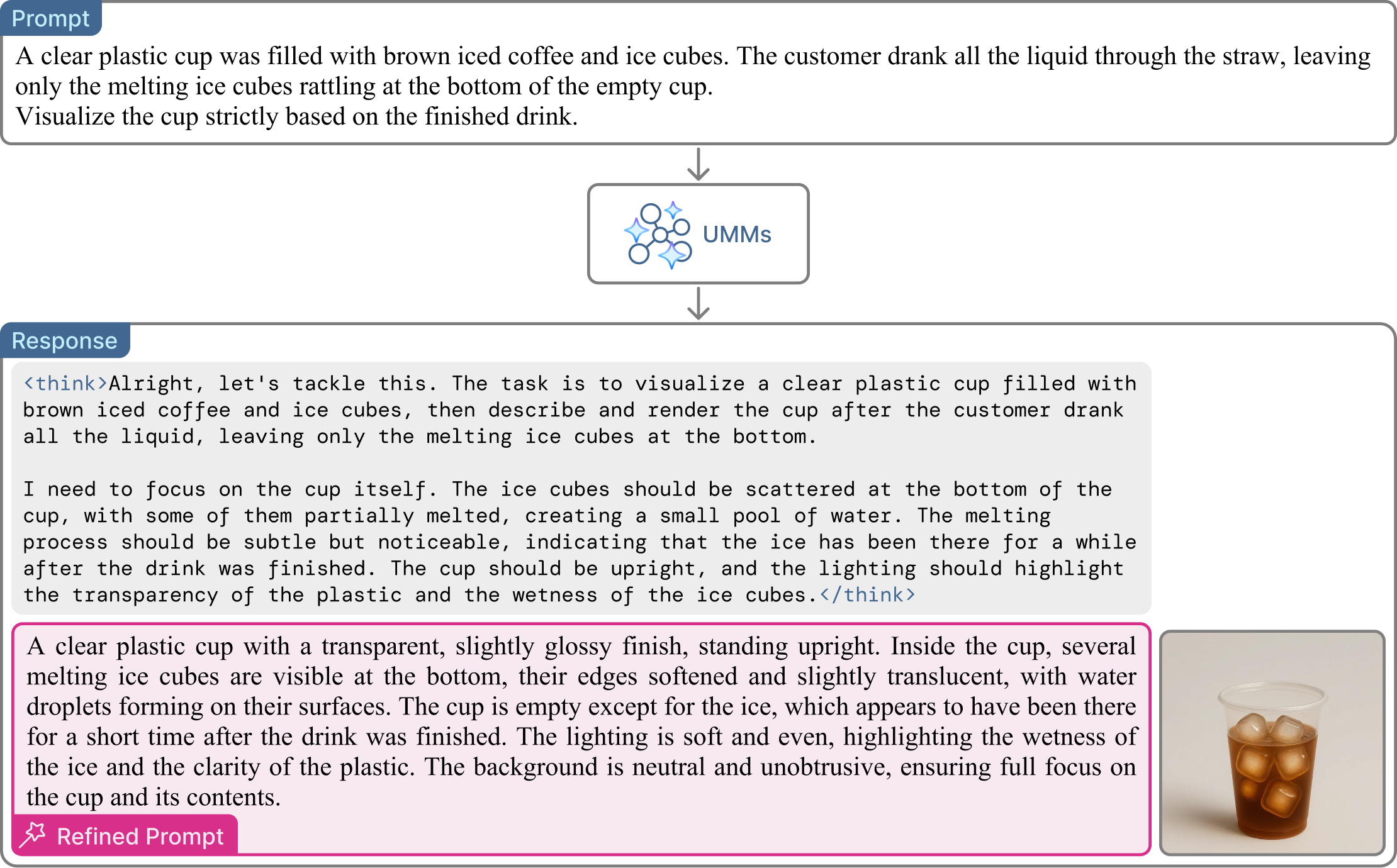}
    \vspace{-1.2em}
    \caption{
\textbf{An Illustrative Example of Task-Specific Error (\textsc{Attribute})}.Despite the prompt explicitly specifying the cup as ``empty except for the ice'', the model generates a brown liquid. This error likely stems from contextual interference, where the model's priors dilute the strict attribute constraint.
    }
    \vspace{-1.2em}
    \label{fig:error_type41}
\end{figure*}

\begin{figure*}[t]
    \vspace{-0.8em}
    \centering
    \includegraphics[width=\textwidth]{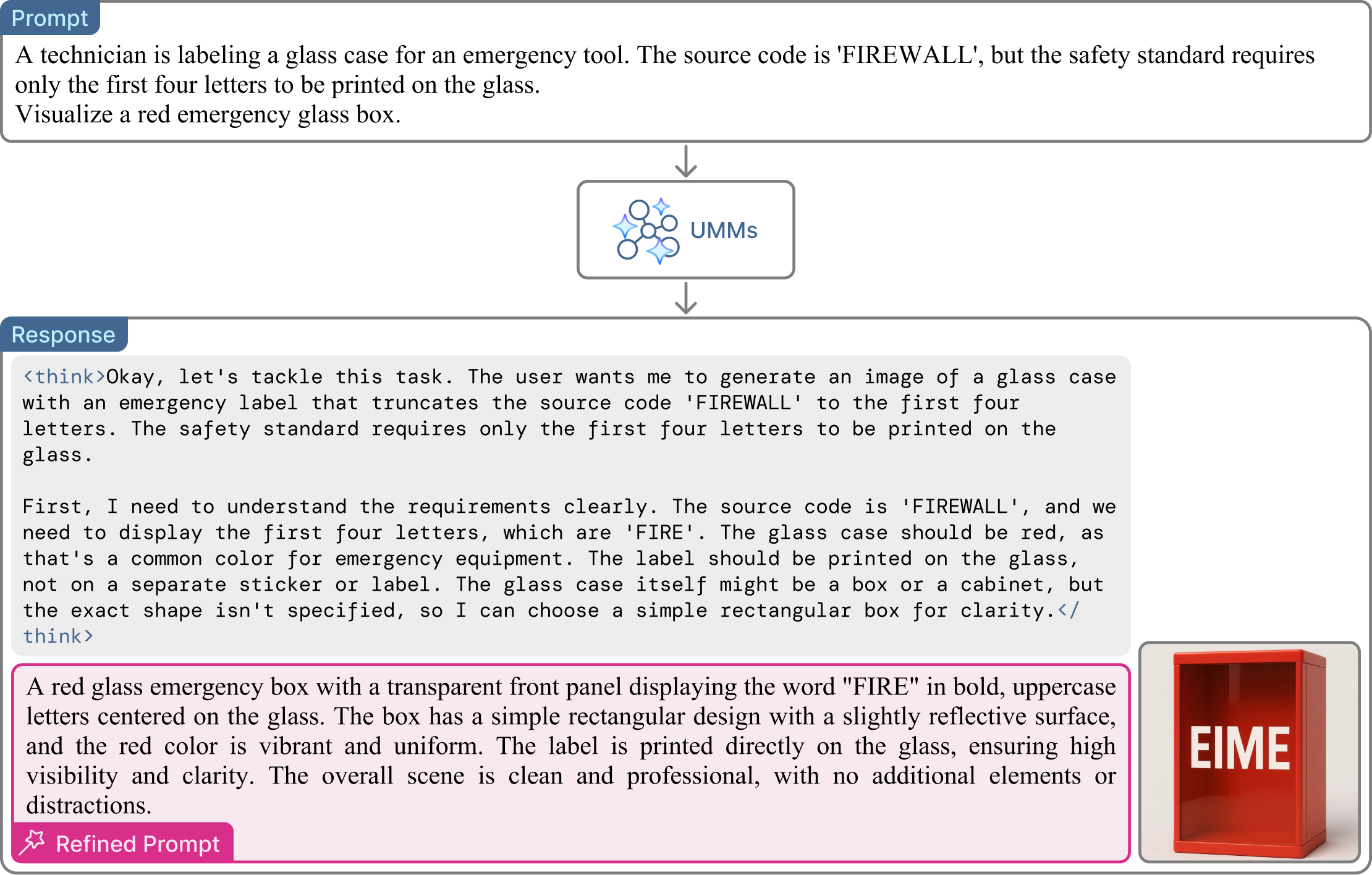}
    \vspace{-1.2em}
    \caption{
\textbf{An Illustrative Example of Task-Specific Error (\textsc{Text}).} The prompt requests the text ``FIRE'', but the model generates ``EIME''. Despite the correct string being present in the refined prompt, the UMM fails to render the characters accurately.
    }
    \vspace{-1.0em}
    \label{fig:error_type44}
\end{figure*}

\paragraph{Task-Specific Errors.} This category accounts for the majority of failures. In these instances, the model successfully avoids the pitfalls of incorrect reasoning, instruction misinterpretation, and concept hallucination, yet still fails to produce a correct output. Crucially, these execution failures occur precisely in the dimensions \bench{} is designed to diagnose. Our fine-grained task design enables analysis of how reasoning affects arithmetic counts, spatial layouts, attribute consistency, and text rendering. This allows researchers to identify specific failure modes in the reasoning-to-generation pipeline. We analyze representative examples across the five tasks below:

\begin{itemize}
\item \textsc{Code}: In Fig.~\ref{fig:error_type42}, the prompt defines a specific HTML table layout for four fashion items. While the model correctly infers the 2$\times$2 grid structure, it fails to map the specific items (Dress, Jeans, Jacket, Sneakers) to their designated table cells, resulting in misalignment despite the structural information being present in the refined prompt. This demonstrates a failure in binding semantic content to structural positions.

\item \textsc{Arithmetic}: As shown in Fig.~\ref{fig:error_type45}, the refined prompt clearly specifies the final state: ``two green apples placed on it and a white bowl containing one green apple.'' The reasoning trace is concise and correct. Nevertheless, the generated image displays three apples on the table and one in the bowl, violating the count constraint. This highlights that even with a correct execution plan, current UMMs struggle to translate precise quantitative specifications into exact object counts, particularly when conditioned on verbose reasoning context.

\item \textsc{Spatial}: In Fig.~\ref{fig:error_type43}, the prompt specifies four quadrants with distinct toppings. While the model generates a pizza, it fails to maintain strict boundary separation and correct topping distribution for each quadrant, instead blending the instructions into a generic pizza image. Examination of the reasoning trace reveals lengthy intermediate steps with detailed visual descriptions. This excessive context likely acts as noise, causing long-context interference that distracts the model from adhering to strict spatial layout constraints.

\item \textsc{Attribute}: As shown in Fig.~\ref{fig:error_type41}, the refined prompt explicitly describes the cup as ``empty except for the ice.'' However, the generated image contains a brown, coffee-like liquid. This error could stem from contextual interference: the reasoning trace explicitly mentions ``brown iced coffee'' to describe the initial state, and this description likely acted as noise, causing the model to erroneously render the initial configuration instead of the final empty state specified in the refined prompt.

\item \textsc{Text}: As seen in Fig.~\ref{fig:error_type44}, the prompt requests the text ``FIRE,'' but the model generates ``EIME''. Despite the refined prompt containing the correct string, the visual generator fails to render the characters accurately. This likely reflects the difficulty of precisely controlling character-level generation when conditioned on verbose reasoning traces, where irrelevant token associations may interfere with accurate text rendering.

\end{itemize}

\subsection{More Qualitative Results}
\begin{figure*}[t]
    \vspace{-0.8em}
    \centering
    \includegraphics[width=\textwidth]{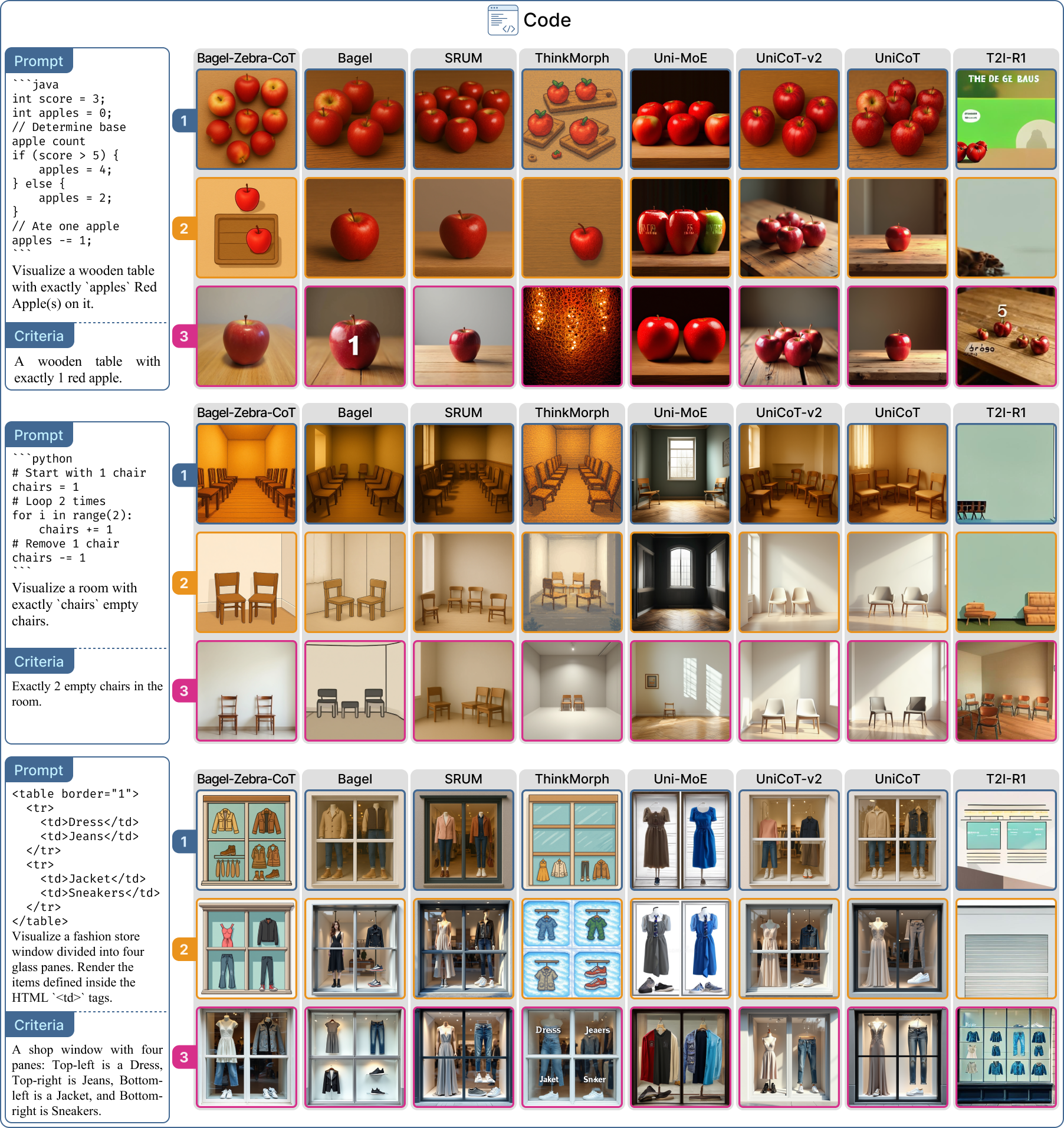}
    \vspace{-1.2em}
    \caption{Qualitative examples for \textsc{Code}.
    }
    \vspace{-1.2em}
    \label{fig:case_study_code}
\end{figure*}

\begin{figure*}[t]
    \vspace{-0.8em}
    \centering
    \includegraphics[width=\textwidth]{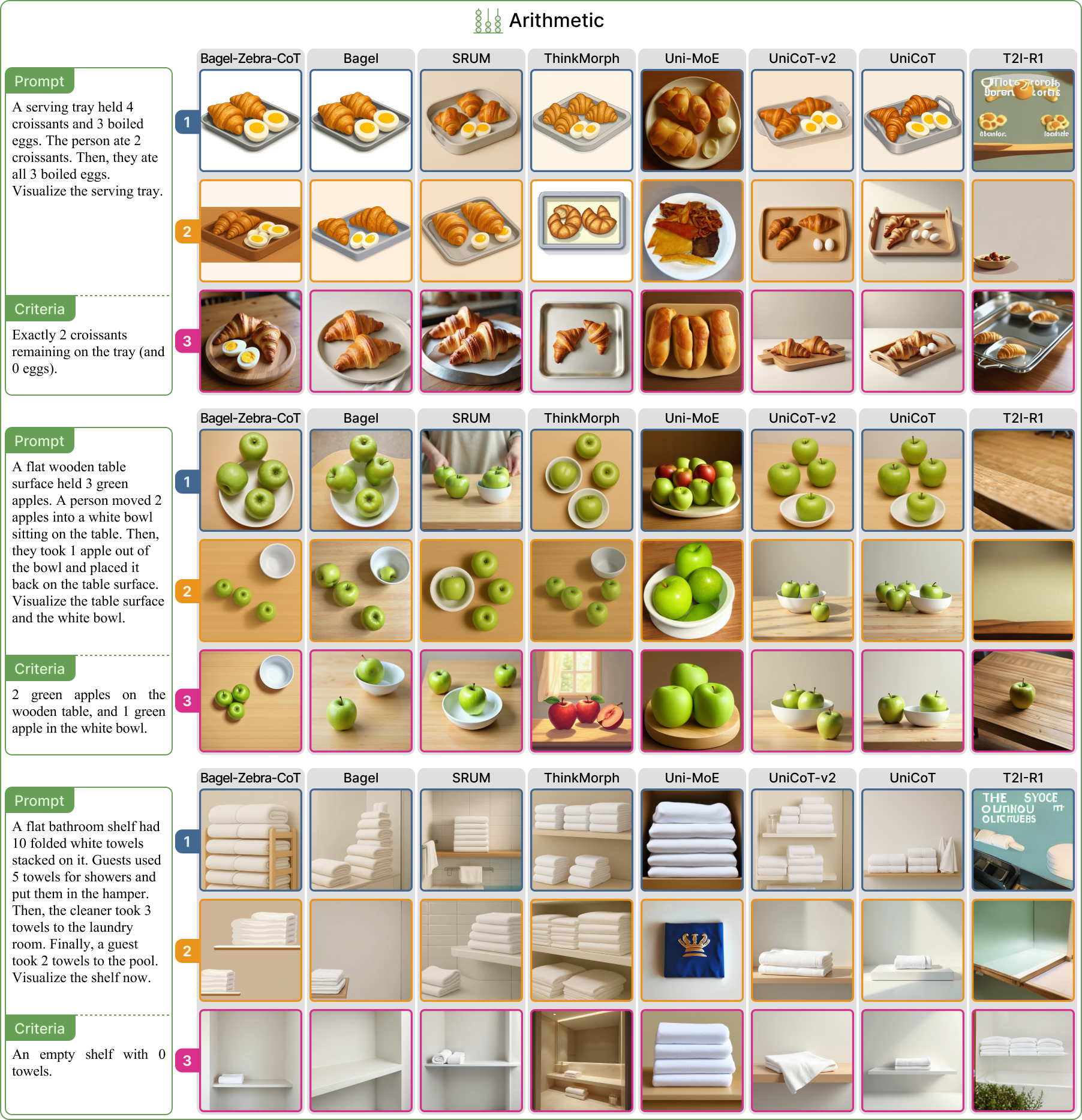}
    \vspace{-1.2em}
    \caption{Qualitative examples for \textsc{Arithmetic}
    }
    \vspace{-1.2em}
    \label{fig:case_study_arithmetic}
\end{figure*}

\begin{figure*}[t]
    \vspace{-0.8em}
    \centering
    \includegraphics[width=\textwidth]{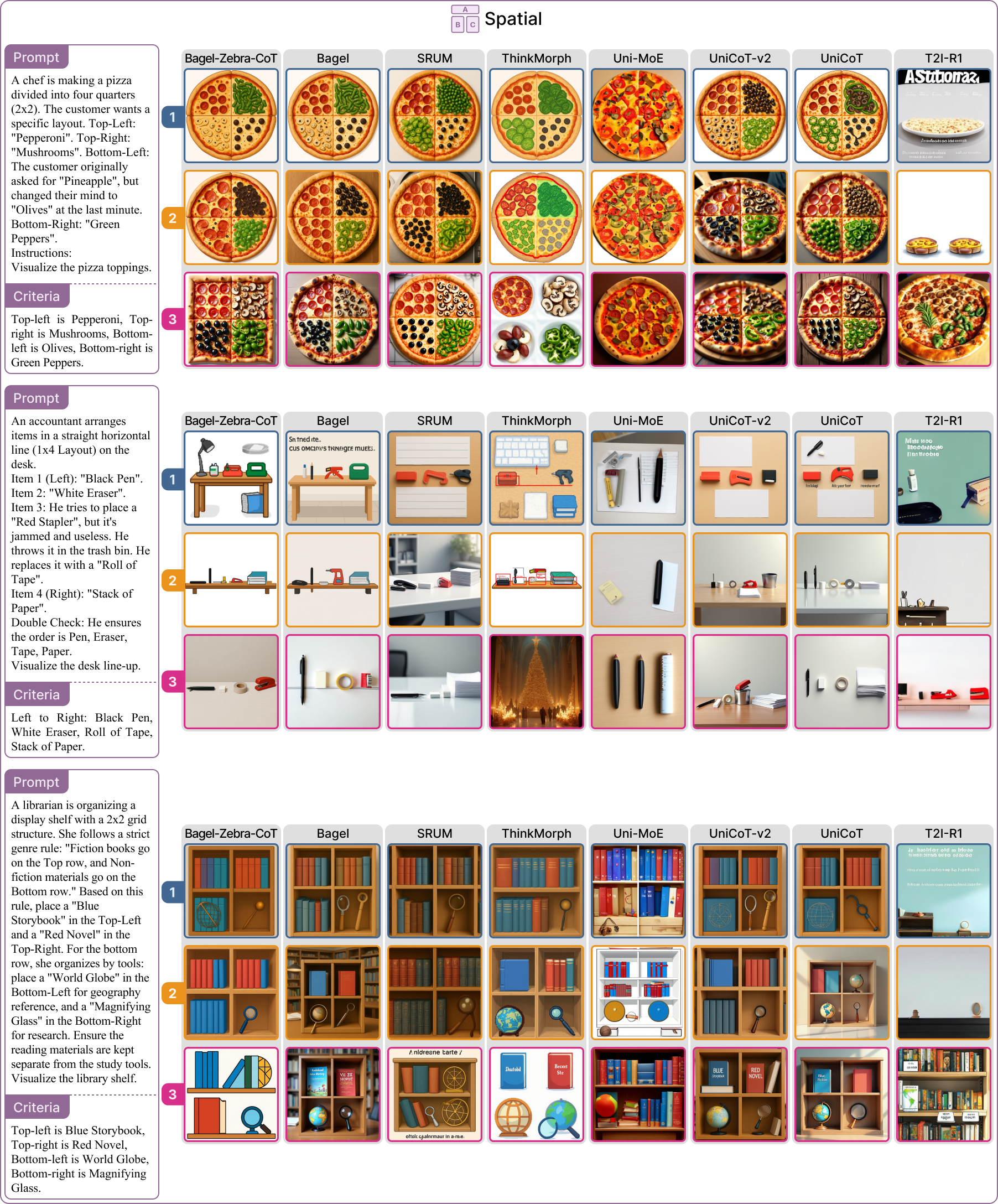}
    \vspace{-1.2em}
    \caption{
    Qualitative examples for \textsc{Spatial}
    }
    \vspace{-1.2em}
    \label{fig:case_study_spatial}
\end{figure*}

\begin{figure*}[t]
    \vspace{-0.8em}
    \centering
    \includegraphics[width=\textwidth]{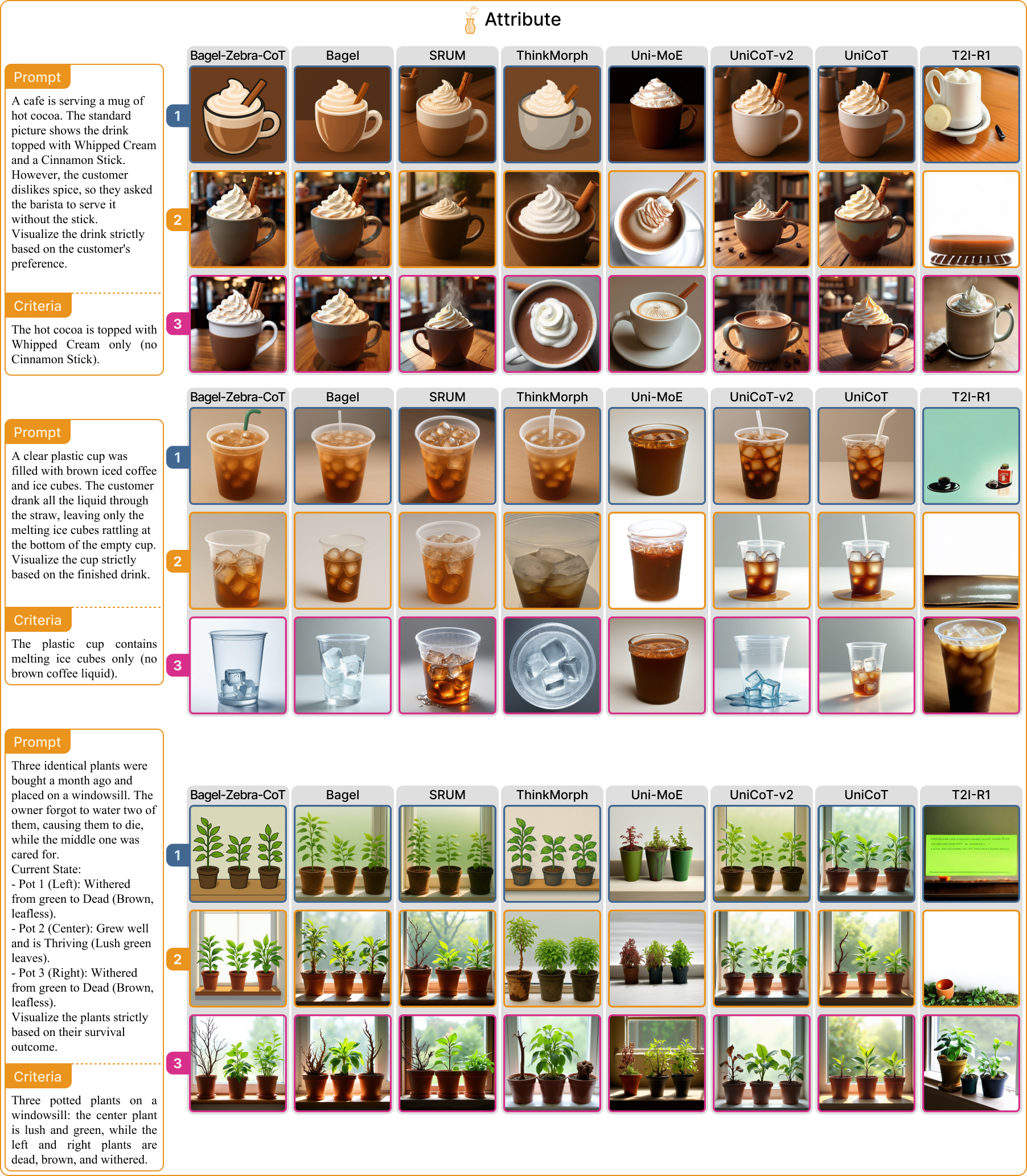}
    \vspace{-1.2em}
    \caption{
    Qualitative examples for \textsc{Attribute}.
    }
    \vspace{-1.2em}
    \label{fig:case_study_attribute}
\end{figure*}

\begin{figure*}[t]
    \vspace{-0.8em}
    \centering
    \includegraphics[width=\textwidth]{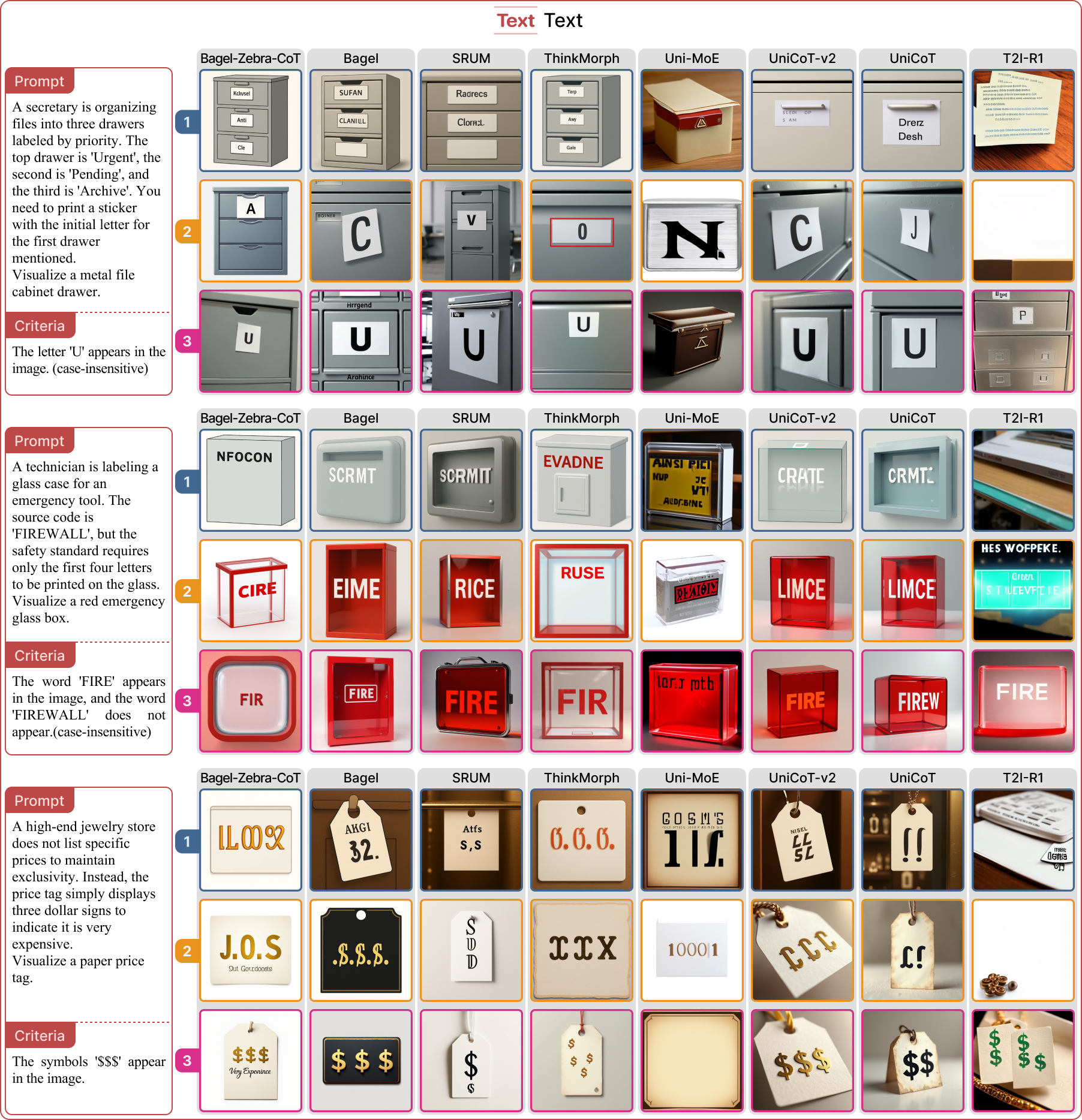}
    \vspace{-1.2em}
    \caption{
    Qualitative examples for \textsc{Text}.
    }
    \label{fig:case_study_text}
\end{figure*}

In addition to the failure modes analyzed above, we provide comprehensive qualitative comparisons across the five tasks defined in \bench{}: \textsc{Code}, \textsc{Arithmetic}, \textsc{Spatial}, \textsc{Attribute}, and \textsc{Text} reasoning. Fig.~\ref{fig:case_study_code} to~\ref{fig:case_study_text} showcase generated samples from representative UMMs across the three evaluation settings.

\section{Ethical Statement}

This work introduces \bench{}, a benchmark for evaluating image generation. Our dataset is human-curated and human-verified, with LLM-assisted candidate generation, and contains only synthetic text prompts and program or code snippets; it does not include personal data, private information, or images of real individuals. The benchmark is designed for capability evaluation rather than deployment. The primary systems evaluated as generation models are publicly available open-source models; closed-source models are used only for data augmentation and evaluation robustness analyses. Its positive impact is to make reasoning-to-generation failures measurable and thereby support the development of more reliable multimodal systems. Potential negative impacts include benchmark gaming, overinterpreting benchmark scores as evidence of general generation safety or capability, and using improved generation fidelity for harmful content creation. Automated evaluators may also inherit model biases despite their high agreement with human judgments. Finally, because the benchmark currently uses English prompts and synthetic scenarios, its concepts may reflect culturally narrow assumptions and should not be treated as globally representative. We will release the data for research use with documentation of its intended diagnostic scope.

\end{document}